\documentclass{article}

\usepackage[preprint]{neurips_2025}

\usepackage{amsmath,amsfonts,bm}

\def\eqref#1{equation~\ref{#1}}

\def\1{\bm{1}}

\DeclareMathAlphabet{\mathsfit}{\encodingdefault}{\sfdefault}{m}{sl}
\SetMathAlphabet{\mathsfit}{bold}{\encodingdefault}{\sfdefault}{bx}{n}

\DeclareMathOperator*{\argmax}{arg\,max}
\DeclareMathOperator*{\argmin}{arg\,min}

\usepackage[utf8]{inputenc} 
\usepackage[T1]{fontenc}    
\usepackage{hyperref}       
\hypersetup{colorlinks=true,linkcolor=red,urlcolor=teal,citecolor=blue,pdfstartview=FitH}
\usepackage{url}            
\usepackage{booktabs}       
\usepackage{amsfonts}       
\usepackage{nicefrac}       
\usepackage{microtype}      
\usepackage{xcolor}         
\usepackage{graphicx}
\usepackage{amsmath}
\usepackage[ruled,vlined]{algorithm2e}
\usepackage{geometry}
\usepackage{multirow}
\usepackage{subcaption}
\usepackage{longtable}
\usepackage[most]{tcolorbox}
\usepackage{caption} 

\def\beqa{\begin{eqnarray}}
\def\eeqa{\end{eqnarray}}
\def\beqann{\begin{eqnarray*}}
\def\eeqann{\end{eqnarray*}}
\def\transpose{{^\top}}

\usepackage{wrapfig}

\newcommand{\capcount}{{\textsc{433}}\xspace}
\newcommand{\taskcount}{{\textsc{11,800 }}\xspace}

\definecolor{promptheaderbg}{RGB}{102,0,102}
\definecolor{promptframebg}{RGB}{239,187,255}

\title{Automated Capability Evaluation of Foundation Models}

\author{
  Arash Afkanpour$^{1}$\thanks{Corresponding author: \texttt{arash.afkanpour@vectorinstitute.ai}} \quad \quad \quad \quad
  Omkar Dige$^{3}$\thanks{Work done while at the Vector Institute} \quad \quad \quad \quad
  Fatemeh Tavakoli$^{1}$ \AND
  Negin Baghbanzadeh$^{1,2}$ \quad
  Farnaz Kohankhaki$^{1}$ \quad
  Elham Dolatabadi$^{1,2}$ \\ \\[4pt]
  $^{1}$Vector Institute, Toronto, Canada \\
  $^{2}$York University, Toronto, Canada \\
  $^{3}$Microsoft
}

\begin{document}

\maketitle

\begin{abstract}
Current evaluation frameworks for foundation models rely heavily on static, manually curated benchmarks, limiting their ability to capture the full breadth of model capabilities. This paper introduces Active learning for Capability Evaluation (ACE), a novel framework for scalable, automated, and fine-grained evaluation of foundation models. ACE leverages the knowledge embedded in powerful frontier models to decompose a domain into semantically meaningful capabilities and generates diverse evaluation tasks, significantly reducing human effort. In Mathematics, ACE generated 433 capabilities and 11,800 tasks, covering 94\% of Wikipedia-defined skills in the domain while introducing novel, coherent ones.
To maximize efficiency, ACE fits a \emph{capability model} in latent semantic space, allowing reliable approximation of a subject model’s performance by evaluating only a subset of capabilities via active learning. It reaches within 0.01 RMSE of exhaustive evaluation by evaluating less than half of capabilities.
Compared to static datasets, ACE provides more balanced coverage and uncovers fine-grained differences that aggregate metrics fail to capture. Our results demonstrate that ACE provides a more complete and informative picture of model capabilities, which is essential for safe and well-informed deployment of foundation models.
\end{abstract}

\section{Introduction}
\label{sec:introduction}

As foundation models grow in scale, generality, and influence across various domains, the challenge of understanding what they can and cannot do becomes increasingly urgent. Capability evaluations serve multiple purposes: they help practitioners select the right model for a given task, guide developers in improving model behavior, and most importantly ensure trustworthiness and safety, particularly in high-stakes domains such as cybersecurity, healthcare, and social engineering. Yet the current evaluation practices are dominated by static, human-curated benchmarks. While useful, these benchmarks quickly fall behind the pace of model development, missing fine-grained skills and introducing costly blind spots \citep{chen2021evaluating,cobbe2021training,dua2019drop,hendrycks2020measuring,hendrycks2021measuring,phan2025humanity,srivastava2022beyond,zellers2019hellaswag,inspect}.

We argue that capability evaluation must itself become adaptive. Instead of freezing tasks in advance, one needs a process that can discover new capabilities as models evolve, generate meaningful and diverse tasks to probe them, and adaptively focus on the most informative regions of the capability space \citet{zhang2024darg,prabhu2024efficient}. Recent advances in frontier and large language models (LLMs) make such adaptivity possible. LLMs can decompose a domain into semantically meaningful capabilities and generate diverse and contextually rich tasks for each capability. However, this power introduces a scalability problem: even a single domain may yield thousands of candidate capabilities, each requiring extensive task sets for reliable scoring. For commercial models (e.g., GPT-4, Claude, Gemini) with usage costs, exhaustive evaluation is prohibitively expensive.

We, therefore, propose to formulate capability evaluation as the problem of \textit{approximating a latent capability function} where the goal is to estimate a model’s competence across a large set of capabilities without exhaustively evaluating every one. The central research question becomes \textit{how to approximate this function effectively when both the number of potential capabilities and the size of task sets required for reliable scoring are large.} Motivated by this question, we present \textbf{A}ctive Learning for \textbf{C}apability \textbf{E}valuation (ACE), a framework for automated, scalable, and fine-grained evaluation of foundation models. ACE operates in two stages:
(1) it uses powerful frontier models to construct structured capability hierarchies and generate tasks with reference solutions; (2) it actively evaluates a subject model by learning its capability function in a latent space and selectively probing informative capabilities. The framework codebase is available at \url{ https://github.com/VectorInstitute/automated_capability_evaluation/} for reproducibility and creating new evaluation benchmarks.

Our contributions are as follows:
\vspace{-2mm}

\begin{itemize}
\itemsep0em 
\item \textbf{Reframing capability evaluation.} We introduce ACE as the first framework that formulates evaluation as \textit{approximating a latent capability function}, rather than exhaustively scoring on fixed benchmarks. We further show that the latent space, constructed via pretrained text encoders and dimensionality reduction, reliably preserves semantic relationships between capabilities, making principled generalization possible.
\item \textbf{Adaptive coverage and efficiency.} By combining LLM-based capability decomposition with active learning in latent semantic space, ACE simultaneously expands coverage (capturing overlooked skills) and improves efficiency (minimizing evaluation cost). This resolves the long-standing trade-off between breadth and scalability in evaluation.
\item \textbf{Large-scale empirical validation.} In Mathematics, spanning \capcount capabilities and \taskcount tasks, ACE reveals capability- and area-level differences invisible to aggregate metrics. It recovers nearly
the entire Wikipedia capability space, showing that
automated benchmarks can surpass static, human-curated ones in coverage, granularity, and cost-effectiveness.
\end{itemize}

\section{Automated Capability Evaluation}
\label{sec:method}

\subsection{Problem Statement}
\label{sec:problem}
In our framework, we define a \emph{capability} as an atomic skill or competence of the subject model (e.g., solving linear equations, factoring polynomials, or summarizing a passage). Capabilities are probed through \emph{tasks}, each of which consists of a problem and a reference solution used for scoring.
We formulate capability evaluation as the problem of approximating a latent \emph{capability function} that reflects how well a model performs across a large space of candidate skills. Following \citet{lu2025automated}, the model under evaluation is referred to as the \emph{subject model}. To construct the capability hierarchy and generate tasks autonomously, our framework relies on a set of frontier models with domain knowledge and reasoning ability. These models, collectively referred to as the \emph{scientist models}, are responsible for proposing candidate capabilities, producing task instances, and supplying reference solutions for evaluation.

Let $\mathcal{C} = \{c_i\}_{i=1}^N$ denote the set of candidate capabilities produced by the scientist models. 
For evaluating a subject model, $\Omega$, each capability $c_i$ can be probed using an evaluation module $\texttt{Evaluate}(\cdot; \Omega)$ that generates multiple tasks, computes their outcomes, and returns an aggregated score $s_i \in \mathbb{R}_+$ for the subject model. Collectively, these scores define the 
latent capability function,
\[
f_\Omega: \mathcal{C} \rightarrow \mathbb{R}_+, \hspace{1mm} \text{where,} \quad f_\Omega(c_i) = \mathbb{E}[\texttt{Evaluate}(c_i;\Omega)]. 
\]

For simplicity, we omit $\Omega$ from the notation when it is clear from context. Obtaining a subject model's capability score via $\texttt{Evaluate}()$ is expensive as it requires designing, solving, and verifying tens of tasks. Therefore, the objective is to approximate $f$ accurately while minimizing the number of calls to 
$\texttt{Evaluate}()$. This differs from static benchmarks, which predefine a fixed subset 
$\{c_i\}_{i=1}^m \subset \mathcal{C}$ and estimate $f$ by exhaustive evaluation on all tasks. Instead, we treat evaluation as an 
\emph{active learning} problem in which we reliably estimate model performance across tasks while minimizing the number of capability evaluations by exploiting semantic relationships across $\mathcal{C}$. 

Two challenges make this problem non-trivial:
\vspace{-2mm}
\begin{itemize}
    \itemsep0em 
    \item \textbf{Coverage}. The space of candidate capabilities is vast and open-ended; without principled generation, important skills may be missed.
    \item \textbf{Efficiency}. Even when capabilities are well-defined, an exhaustive evaluation could be expensive. A scalable framework must identify a small, informative subset that suffices to approximate the capability function reliably. 
\end{itemize}

ACE addresses these challenges through two components: structured capability discovery, which organizes $\mathcal{C}$ into meaningful hierarchies, and latent modeling with active learning, which adaptively approximates $f$.

\subsection{Capability Hierarchy and Task Design}
\label{sec:design}

\begin{figure}
    \centering
    \includegraphics[width=\textwidth]{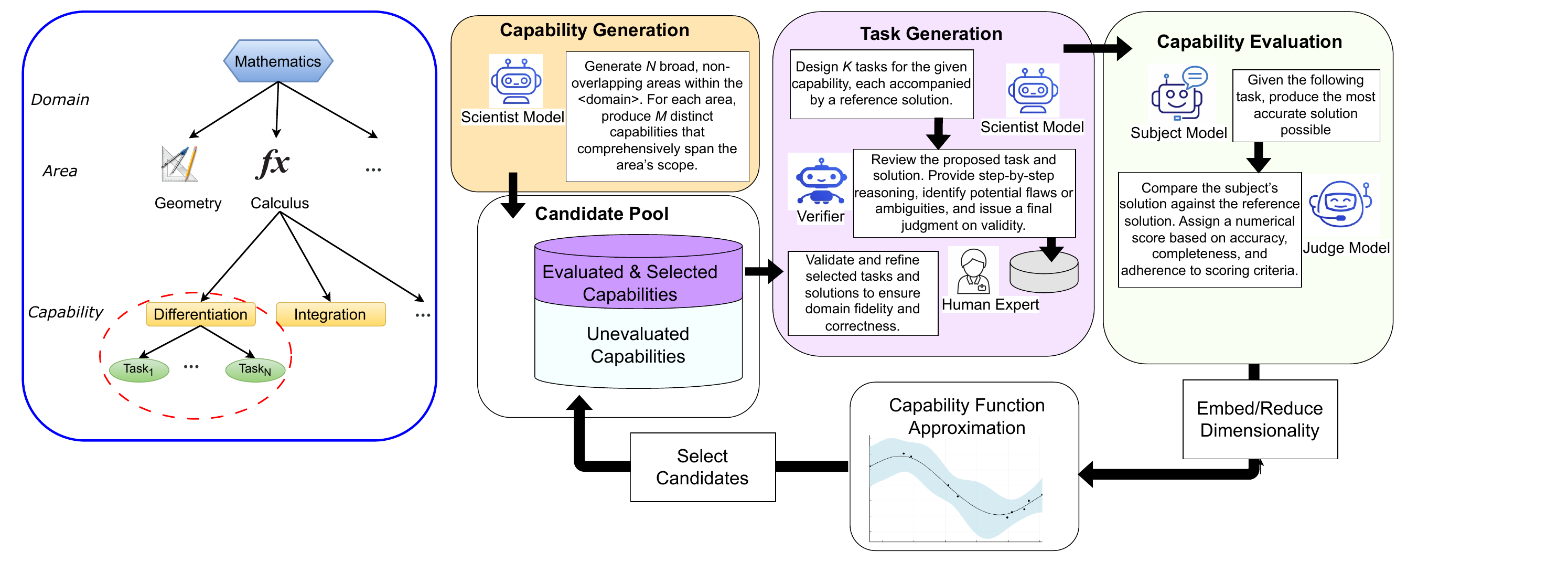}
    \caption{An overview of ACE. \textbf{Left:} Example capability hierarchy in Mathematics. \textbf{Right:} The ACE pipeline combining automated capability generation, task generation and verification, and active learning in latent space for efficient model evaluation.}
    \label{fig:hierarchy_and_ace_pipeline}
\end{figure}

\textbf{Capability Hierarchy.} Building on the  definition of capabilities above, we next describe how they are organized and operationalized. At the top level, a domain, e.g., Mathematics, is divided into broad \emph{areas} such as Algebra, Calculus, or Probability and Statistics; each area is then refined into specific \emph{capabilities}, for example, Probability and Statistics is further broken down into capabilities such as Bayesian Inference, Markov Chain Probabilities, etc. This hierarchy is extensible and can support multiple levels of granularity depending on evaluation needs (Figure~\ref{fig:hierarchy_and_ace_pipeline}). 

\textbf{Task Instantiation.} For each capability, the scientist models generate a set of \emph{tasks}, each consisting of a problem and reference solution. Task formats are domain-dependent: in structured domains like Mathematics, problems usually admit unique solutions that can be deterministically verified; in open-ended domains such as summarization or scientific writing, multiple valid responses may exist, requiring more nuanced evaluation.

\textbf{Scoring.} The performance of a subject model on a task can be quantified either as a binary score, e.g., solved (1) vs. not solved (0), in domains with well-defined solutions, or as a continuous value in $[0,1]$ to capture partial correctness or graded quality in open-ended domains. To obtain reliable capability-level estimates, the subject model is evaluated on a sufficiently large set of tasks through \texttt{Evaluate()}, which computes and aggregates task-level scores. By default, the mean is used; when tasks vary in difficulty or importance, weighted averages are applied, and in settings sensitive to outliers, the median is preferred.

\textbf{Verification.} Since the problem and reference solution for each task are generated automatically by the scientist models, we introduce a verification step. First, verification models review reference solutions for correctness. To further safeguard quality, we conduct targeted human inspection of outputs from both the task generation and verification stages (Appendix~\ref{sec:manual_inspection}). This ensures that the ground truth used for evaluation is reliable and reduces the risk of propagating errors during scoring. Second, subject model responses are evaluated against these references through \texttt{Evaluate()}. For structured tasks with deterministic answers (e.g., Mathematics), correctness is established through direct comparison. For close-ended solutions, exact match will be considered for evaluation. For open-ended tasks, we employ a judge model that scores responses against the task and reference solution on criteria such as accuracy, completeness, coherence, and relevance. 

The judge model
is provided with the task description, the reference solution, and the subject model’s response. Judge prompts can be calibrated, and multiple judges can be ensembled to improve robustness. This layered approach ensures that both the ground-truth references and the subject model’s outputs are evaluated rigorously and consistently. An abstract overview of our pipeline is provided in Figure~\ref{fig:hierarchy_and_ace_pipeline} (Right).

\subsection{Latent Modeling of Capabilities}
\label{sec:latent_modeling}

\textbf{Embedding.} We assume that capabilities in a domain are specified in a discrete space $\mathcal{T}$. For example, $\mathcal{T}$ could be the text space, where each capability is described by a short natural language statement.
Direct function approximation in this space is challenging. We, therefore embed capabilities into a continuous latent space $\mathcal{Z} \subset \mathbb{R}^d$ using a pretrained encoder $E: \mathcal{T} \rightarrow \mathcal{Z}$. Each capability $t_i \in \mathcal{T}$ is mapped to $z_i = E(t_i)$, yielding dense semantic representations that support generalization across related capabilities. We assume the underlying capability function, $f$, is smooth in this space. This assumption is supported by empirical observations that LLMs exhibit correlated performance across related skills \citep{wang2024fac,siska2024examining,ilic2024evidence}.

A key requirement of our approach is that semantically similar capabilities in $\mathcal{T}$ are mapped to nearby points in $\mathcal{Z}$. This property is essential for reliable generalization and uncertainty modeling. In Section~\ref{sec:semantic_relationship}, we empirically demonstrate that modern encoders satisfy this condition. Given an initial set of capability-score pairs $\{(t_i, s_i)\}_{i=1}^N$, with scores $s_i$ obtained from the subject model, the learning task reduces to approximating the capability function $f$ from the set of embedded pairs $\{(z_i, s_i)\}_{i=1}^N$.

\textbf{Function Approximation via Active Learning.} Exhaustive evaluation of all capabilities in $\mathcal{C}$ could be very expensive, as each call to \texttt{Evaluate()} involves generating, verifying, and scoring many tasks. To address this, we employ \emph{active learning} to adaptively select informative capabilities. At each iteration, we compute the active learning \emph{acquisition} scores\footnote{Acquisition function in active learning refers to the function used for selecting a candidate in each round.} across unevaluated capabilities, select the optimal candidate, invoke \texttt{Evaluate()} on it, and add it to the training set for approximating the capability function. We then update the regression model. For regression, we adopt Gaussian Processes (GP), which provide both predictive means and  uncertainty estimates, making it suitable for active learning \citep{malkomes2019automating,gorissen2009automatic,fu2022active,riis2021active,chabanet2021coupling}. In our implementation we adopt the variance-reduction strategy of \citet{cohn1996neural}, which selects the candidate capability that yields the largest expected reduction in posterior variance over the domain of $f$. This choice offers strong sample efficiency when evaluation budget is limited. Further details on GP regression and acquisition alternatives are provided in Appendix~\ref{app:active_learning_gp}.

\textbf{Dimensionality Reduction.} Capability embeddings are often high-dimensional, which, due to curse of dimensionality, can hinder GP regression. To address this, we apply dimensionality reduction $
\varphi : \mathbb{R}^d \to \mathbb{R}^{d'} \quad \text{where } d' \ll d$, using methods such as Principal Component Analysis (PCA) or t-SNE \citep{van2008visualizing}.

Bringing everything together, Algorithm~\ref{alg:ace} in Appendix~\ref{app:alg} presents the full active learning procedure for capability function approximation.

\section{Experiments}
\label{sec:experiments}

\subsection{Setup}
\label{sec:setup}
 
To assess the ACE framework, we focus on the domain of mathematics, which offers a hierarchical structure and well-defined problem formats, making it a natural testbed for capability-centric evaluation.
In our experiments, we employ two scientist models, OpenAI $\texttt{gpt-4o}$\footnote{\url{https://platform.openai.com/docs/models/gpt-4o}} and Anthropic \texttt{Claude 3.7 Sonnet}\footnote{\url{https://www.anthropic.com/news/claude-3.7-sonnet}}.\footnote{At the time of this analysis these models were the frontier models of these companies.} The capability hierarchy is generated by the \texttt{gpt-4o} model, while tasks were generated by both models.
To construct a diverse capability set, the scientist model is first prompted to propose broad and distinct areas within Mathematics. For each area, it is further prompted to produce specific capabilities in the modified \texttt{METR}\footnote{\url{https://metr.org/}} format following \citet{lu2025automated}. Each capability includes a \texttt{name}, \texttt{description}, and a corresponding \texttt{Python class}, which specifies exemplar tasks, task-solving instructions, and the scoring method. Full prompts for area- and capability-level generation are provided in Appendix~\ref{sec:area_gen_prompts} and~\ref{sec:cap_gen_prompts}.

These capabilities serve as input to the task generation pipeline. The pipeline begins by generating multiple diverse problems per capability. Each problem is then solved using the task-solving instructions specified in the capability's \texttt{Python class} to produce a solution. Together, the problem and solution form a complete task. We verify each task using a separate LLM call to confirm the correctness of the solution and to filter out incorrectly solved or unsolved tasks. Full task generation prompts are provided in Appendix~\ref{sec:task_gen_prompts}.

For evaluation, we use the \texttt{Inspect} framework \citep{inspect}, which dynamically generates evaluation scripts based on the task-solving instructions and scoring method defined in each capability's \texttt{Python class}. For Mathematics, we adopt a binary scoring scheme: the subject model receives a score of 1 if its solution matches the reference solution, and 0 otherwise.

Following this procedure, we generate a benchmark of \capcount distinct capabilities spanning 10 diverse mathematical areas.  
Although ACE does not require pre-generated task sets, for the purposes of experimentation and analysis we generated \taskcount tasks, with 27 tasks per capability on average. Experiments proceed in four stages: (i) coverage and task validity, (ii) capability benchmarking, (iii) validation of latent-space structure, and (iv) adaptive evaluation that approximates the capability function with active learning.

\subsection{Coverage and Task Validity}
Our first question is \textit{whether ACE-generated benchmarks provide comprehensive coverage and valid, discriminative tasks compared to established resources?} We compare three capability sets: (i) 287 ground-truth capabilities from Wikipedia (all sub-areas of mathematics from Wikipedia\footnote{\url{https://en.wikipedia.org/wiki/Glossary_of_areas_of_mathematics}}), (ii) our ACE-generated synthetic benchmark, and (iii) Static human-curated math datasets, including MATH~\citep{hendrycks2021measuring} and GSM8K~\citep{cobbe2021training}.

\textbf{Wikipedia Coverage.} To quantify the overlap and differences between Wikipedia and ACE capability sets, we perform bidirectional matching analyses. Specifically, we use \texttt{Qwen2.5-32B-Instruct} with a classification prompt to map each capability from a source set to the most relevant capability in the target set. This constitutes a many-to-one matching problem: a given capability may map to a single best counterpart in the target set, but multiple source capabilities may map to the same target. Since many-to-one mappings are not symmetric, we conduct the analysis in both directions: Wikipedia $\rightarrow$ ACE and ACE $\rightarrow$ Wikipedia. Here, A $\rightarrow$ B indicates how many of capabilities in A are covered by B.

From the perspective of Wikipedia $\rightarrow$ ACE, 269 of 287 Wikipedia capabilities (94\%) were matched to ACE capabilities, suggesting that ACE reliably captures nearly all widely recognized mathematical skills. From the perspective of ACE $\rightarrow$ Wikipedia, 405 of 433 ACE capabilities (93\%) were covered by Wikipedia, while the remaining 28 appear to represent novel and potentially meaningful capabilities not explicitly covered in Wikipedia.

\begin{figure}
    \centering
    \begin{subfigure}{0.6\textwidth}
        \centering
        \includegraphics[width=\textwidth]{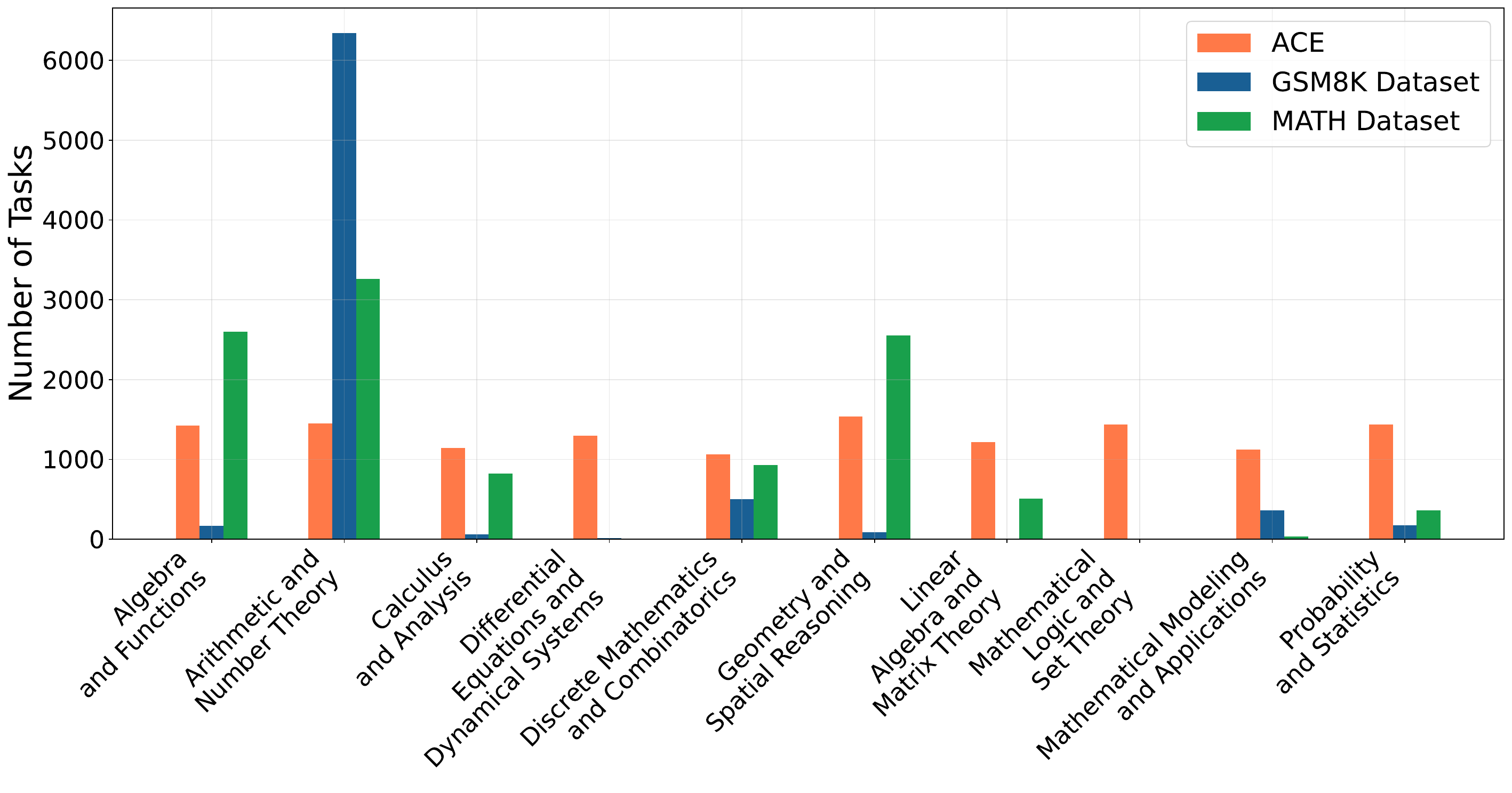}
        \caption{}
        \label{fig:hierarchy}
    \end{subfigure}
    \begin{subfigure}{0.38\textwidth}
        \centering
        \includegraphics[width=\textwidth]{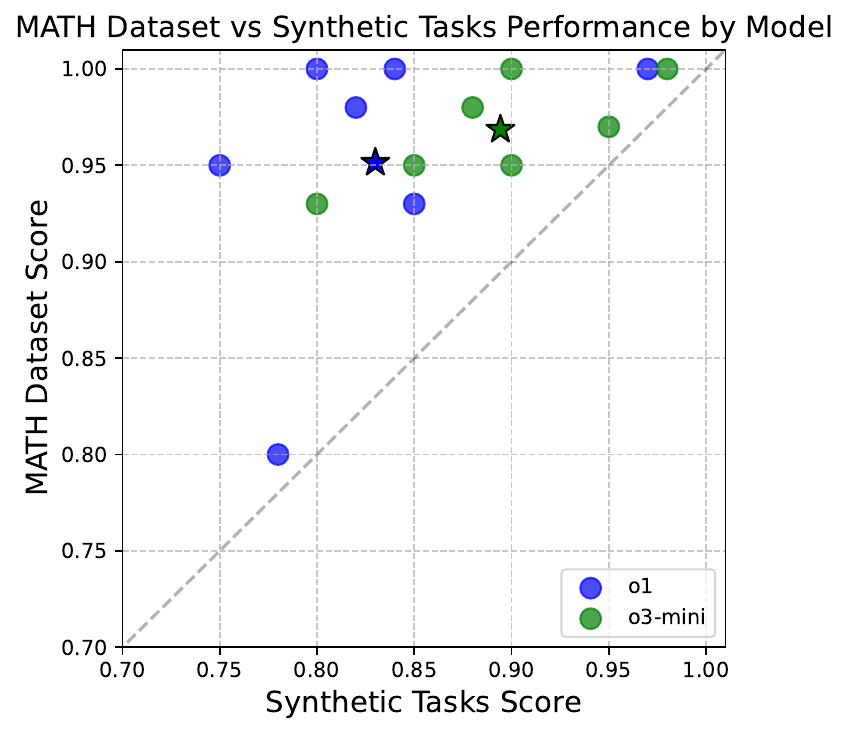}
        \caption{}
        \label{fig:static_vs_synthetic}
    \end{subfigure}
    \caption{Coverage and validity of ACE-generated benchmarks. (a) Task distributions across mathematical areas for ACE (orange), GSM8K dataset (blue) and MATH dataset (green). (b) Subject model performance on MATH vs. ACE (synthetic) tasks. Stars indicate average score across all capabilities. }
    \label{fig:fig2}
\end{figure}

\textbf{Dataset Coverage.} To assess the coverage of static versus synthetic benchmarks, we categorize problems from MATH and GSM8K datasets into the high-level mathematical areas defined in ACE (see Section~\ref{sec:static_dataset_categorization} for details). The resulting distributions are shown in Figure~\ref{fig:fig2}(a). 
GSM8K exhibits a highly skewed distribution with a large fraction of tasks falling into the Arithmetic and Number Theory area, while other important areas (e.g., Differential Equations, Discrete Mathematics) are scarcely represented or entirely absent. The MATH dataset is less skewed than GSM8K, yet it lacks coverage in areas like Differential Equations. In contrast, ACE tasks are generated to achieve balanced coverage across all areas by design. This comparison highlights a key limitation of static benchmarks such as GSM8K and MATH: their task distributions often reflect dataset construction biases, leading to overrepresentation of certain skills and underrepresentation of others. Synthetic benchmarks like ACE mitigate this issue by enabling systematic and uniform coverage across the full capability space of the domain.

\textbf{Discriminative Power of Synthetic Benchmarks.} In this study we compare subject model scores on a synthetic benchmark and the MATH dataset. To construct the synthetic benchmark for each of the seven areas in MATH\footnote{Pre-algebra, Algebra,
Number Theory, Counting and Probability, Geometry, Intermediate Algebra, and Pre-calculus}, we generate  tasks using the scientist model. For each area, we then evaluate the performance of a subject model on synthetic tasks and the corresponding subset of MATH problems. We evaluate two subject models and show the results in Figure~\ref{fig:fig2}(b). 
For both subject models, comparing the distribution of scores across the two benchmarks
 reveals greater variation in scores on the synthetic benchmark. This indicates that
 tasks in the synthetic benchmark span a broader range of problem types and difficulties for each area.
Consequently, this benchmark provides a more nuanced and discriminative assessment of model
strengths and weaknesses.

\subsection{Fine-Grained Benchmarking}
\label{sec:benchmarking-llms}

\begin{figure}
    \centering
    \begin{subfigure}[c]{0.6\textwidth}
        \centering
        \includegraphics[width=0.83\textwidth]{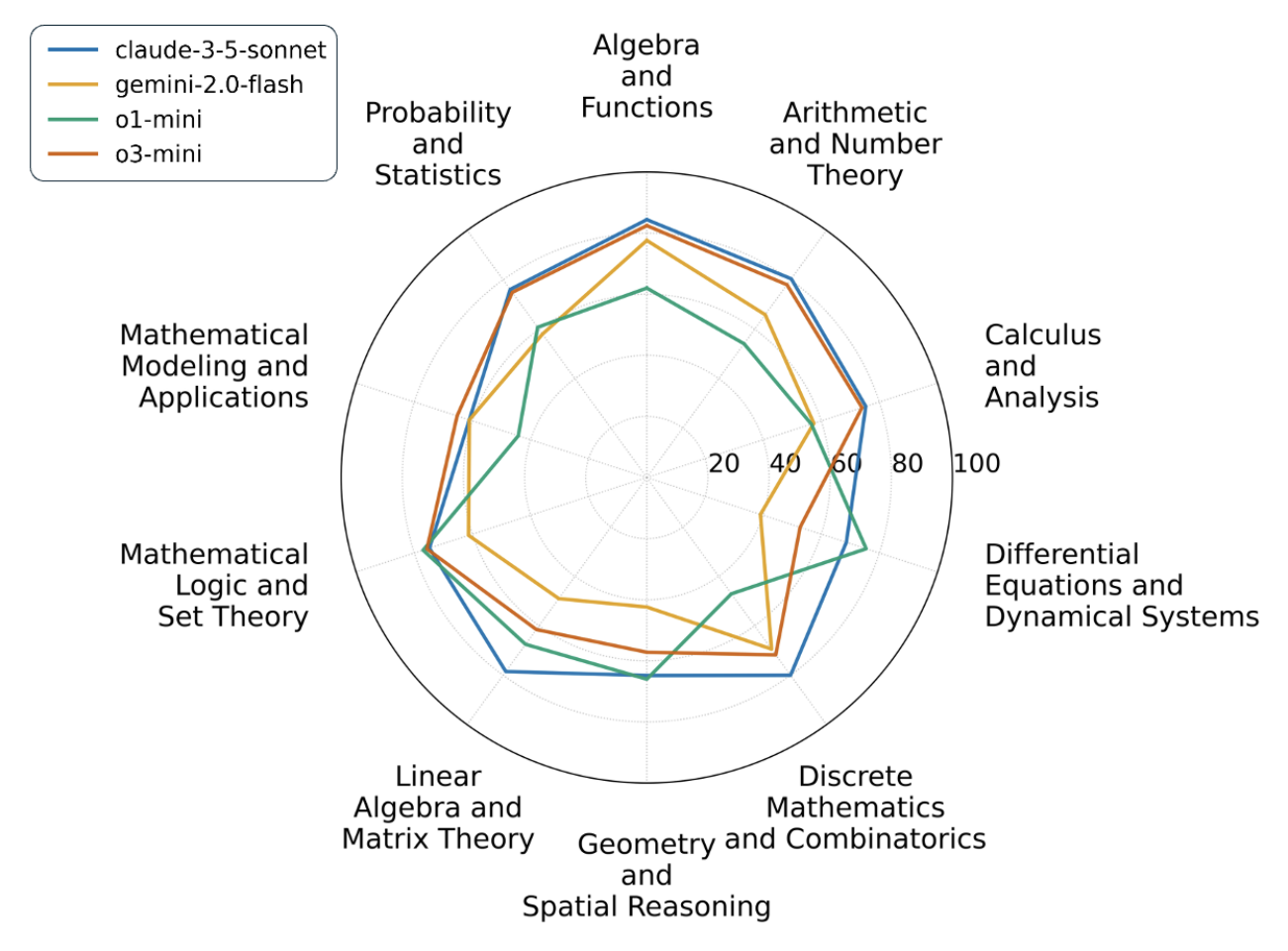}
        \caption{}
        \label{fig:main_spider_chart}
    \end{subfigure}
    \begin{subfigure}[c]{0.36\textwidth}
        \centering
        \includegraphics[width=\textwidth]{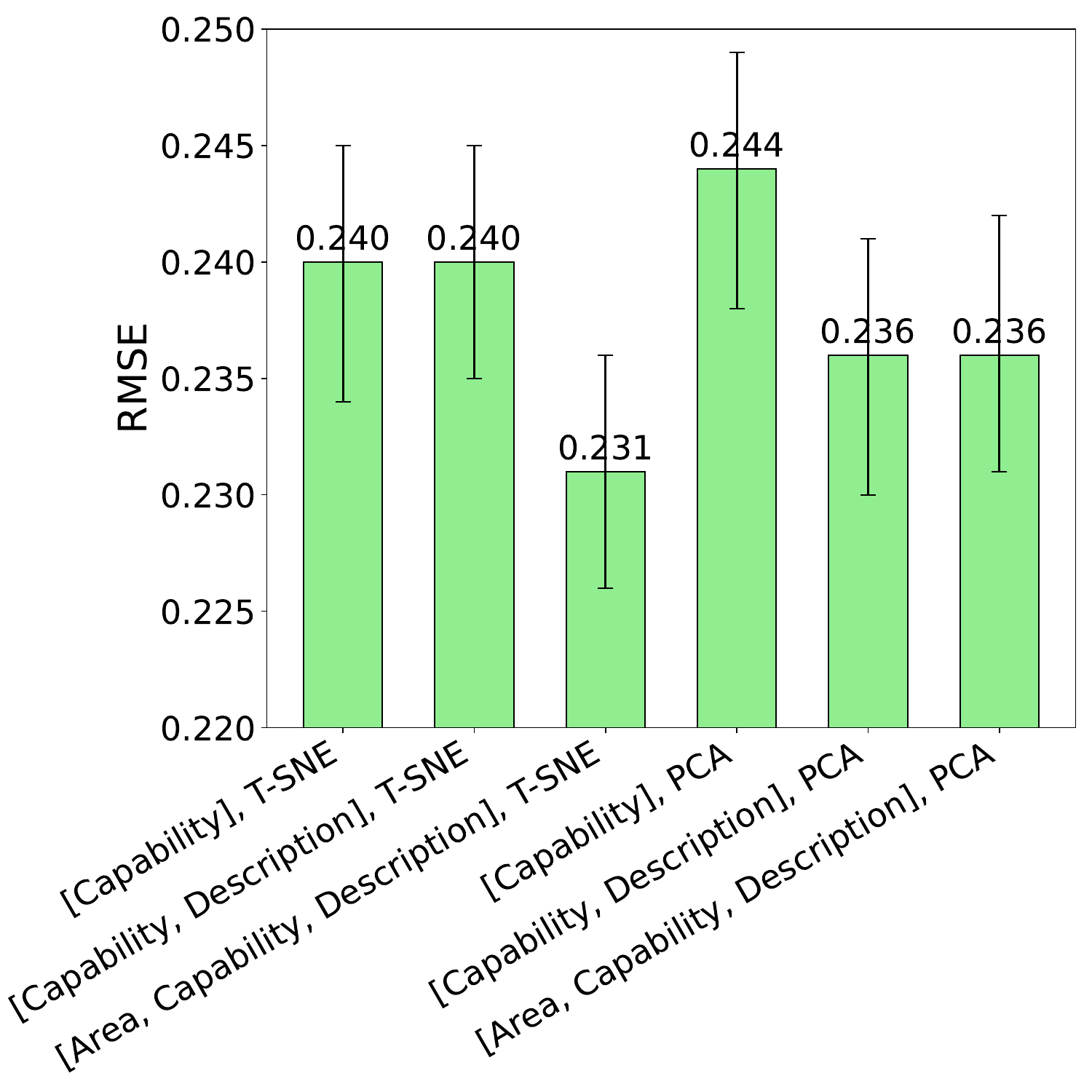}
        \caption{}
        \label{fig:score_spider}
    \end{subfigure}
    \caption{(a) Area-level benchmarking: subject model scores across mathematical areas. The reported score for each area is the average score of all capabilities within that area. (b) Semantic structure in latent space: Effect of input text and dimensionality reduction technique on capability function approximation.}
    \label{fig:math_capability_performance}
\end{figure}

Using ACE, we perform fine-grained evaluation of four  subject models on all \capcount  capabilities of Mathematics. Area-level scores are computed by averaging capability-level scores. Figure~\ref{fig:math_capability_performance}(a) shows the results. Among these subject models \texttt{Claude-3.5-Sonnet} is the strongest and most consistent performer, maintaining high accuracy across nearly all areas. \texttt{o3-mini} follows closely. \texttt{o1-mini} performs well in Differential Equations and 
Dynamical Systems, but lags behind in several other areas. Finally, \texttt{Gemini-2.0-Flash} exhibits relatively low performance in areas such as Differential Equations or Calculus. These results illustrate the value of a structured fine-grained evaluation: even among generally
strong models, there are differences in performance  that may not be apparent in aggregate performance metrics.

\subsection{Semantic Structure in Latent Space}
\label{sec:semantic_relationship}
Reliable approximation of the capability function, $f$, depends on whether the latent space $\mathcal{Z}$
preserves semantic relationships between capabilities. In particular, capabilities within the same area
should be embedded close to each other in $\mathcal{Z}$. Two components influence the structure of the latent space: the text encoder,
which maps natural language descriptions of capabilities to high-dimensional embeddings, and the
dimensionality reduction technique used to project these embeddings into a lower-dimensional space.

We first study the effect of the text encoder in isolation. We embed a subset of 20 capabilities sampled from 5 areas of Mathematics using the OpenAI \texttt{text-embedding-3-small} model\footnote{\url{https://platform.openai.com/docs/guides/embeddings/}}(512-dimensional output). Each embedding is generated by concatenating the \texttt{area name}, \texttt{capability name}, and \texttt{capability description}. Pairwise cosine similarity analysis reveals clear intra-area clustering (Appendix~\ref{app:capability_embedding_heatmap}), indicating that embeddings capture meaningful semantic relations.

Next, we assess the combined effect of the text encoder and dimensionality reduction. We embed all
\capcount capabilities using the same encoder and project the resulting representations into a 2D latent space using t-SNE or PCA. Figure~\ref{fig:capability_distribution} in the Appendix shows the distribution of capabilities in the latent space.
Both techniques preserve
semantic relationships to varying degrees, but t-SNE produces more distinct clusters for capabilities within each area.

Finally, we assess how the encoder input choice and dimensionality reduction affect approximation of the capability function. Options for encoding a capability include (1) capability name, (2) capability name and description, and (3) capability name, description, and area name. Dimensionality reduction techniques consist of t-SNE and PCA. A Gaussian Process model is trained on 80\% of the capability set, and Root Mean Square Error (RMSE) is reported on the test set. Figure \ref{fig:math_capability_performance}(b) summarizes the results. We find that including richer input text (capability name, description, area name) and t-SNE yields the best performance.

\subsection{Adaptive Evaluation for Efficient Approximation}
\label{sec:al_ace}
\begin{figure}
    \centering
    \begin{subfigure}[c]{0.49\textwidth}
        \centering
        \includegraphics[width=\textwidth]{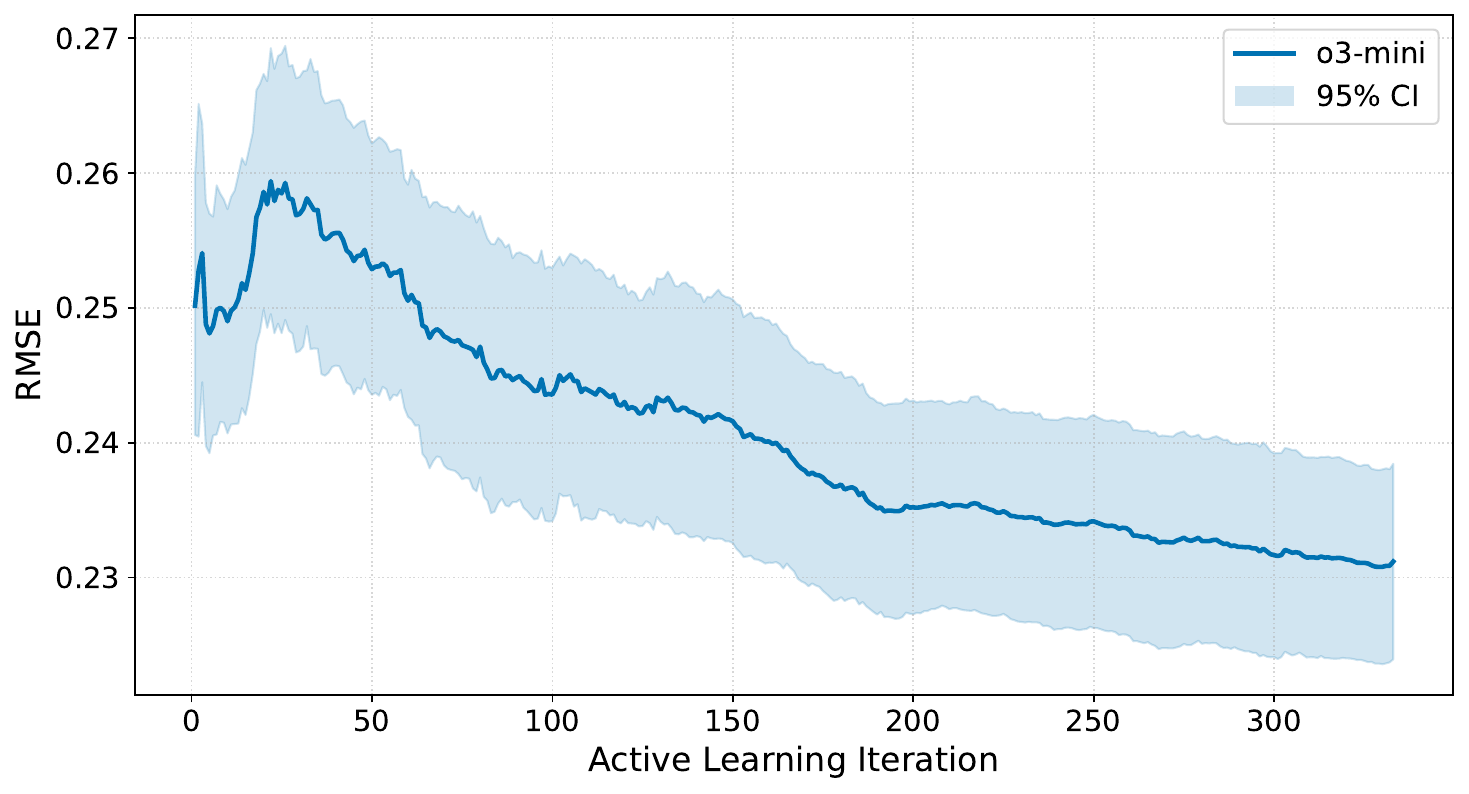}
    \end{subfigure}
    \begin{subfigure}[c]{0.49\textwidth}
        \centering
        \includegraphics[width=\textwidth]{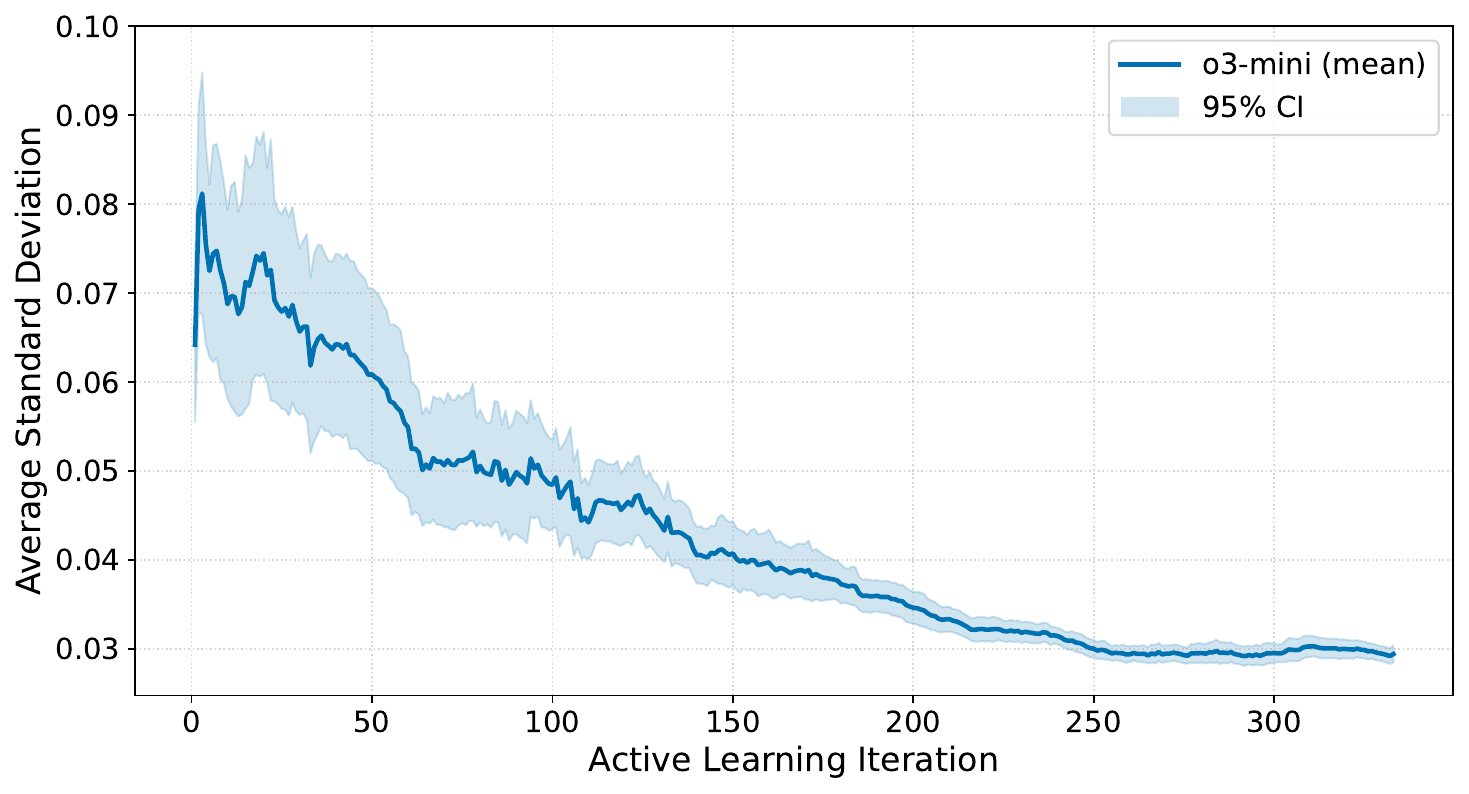}
    \end{subfigure}
    \caption{Performance of approximating the capability function. (Left) RMSE, (Right) Uncertainty (average standard deviation) over iterations of active learning. Shaded areas indicate 95\% confidence intervals.}
    \label{fig:al_results}
\end{figure}

We conduct an ablation study of active learning in ACE to investigate the trade-off between efficiency and accuracy in function approximation. In practical model evaluation, the candidate pool for active learning would consist of all capabilities. For this experiment, however, we allocate 80\% of the capabilities to the candidate pool and reserve the remaining 20\% as a held-out test set for measuring generalization error. A GP model is initialized with two randomly selected capabilities from the training set and iteratively updated through active learning. At each iteration, the capability chosen is the one that yields the largest expected reduction in posterior variance over the domain of $f$ \citep{cohn1996neural} (see Appendix~\ref{app:active_learning_gp}). We use the scores of the \texttt{o3-mini} subject model for this study.
Figure~\ref{fig:al_results} presents RMSE (left), and predictive uncertainty (right) on the test set across active learning iterations.  These results indicate that by evaluating the subject model on fewer than 50\% of the capabilities (150 out of 346), the GP model achieves an RMSE within 0.01 of the target generalization error. Moreover, we observe a steady reduction in predictive uncertainty throughout the process.\footnote{Additionally, Figure~\ref{fig:al_spider_chart} in the Appendix shows area-level scores of the subject model when the capability function is fit on a fraction of capability scores. Note that these are reported scores only on the test set.} These findings demonstrate that incorporating active learning in ACE provides effective generalization while substantially reducing evaluation cost.

\section{Related Work}

\textbf{Traditional Evaluation.} Early evaluation efforts relied on static, manually curated benchmarks such as MMLU~\citep{hendrycks2021measuringmassivemultitasklanguage}, BIG-bench~\citep{srivastava2022beyond}, and HELM~\citep{liang2022holistic}, which aimed for broad coverage of general knowledge and reasoning. Other datasets target specific weaknesses, e.g., TruthfulQA~\citep{lin2022truthfulqameasuringmodelsmimic} for factual reliability and ARC~\citep{clark2018think} for scientific reasoning. While influential, these benchmarks are inherently static, susceptible to contamination~\citep{deng2024investigatingdatacontaminationmodern}, and uneven across domains. Mathematics, for example, is relatively well served (e.g., GSM8K~\citep{cobbe2021training}, MATH~\citep{hendrycks2021measuring}, but many applied and professional areas lack dedicated benchmarks. This motivates the need for adaptive frameworks that go beyond frozen datasets.

\textbf{Automated Evaluation.} Recent work leverages LLMs to generate, adapt, or score test cases, aiming to scale evaluation beyond fixed datasets. Model-assisted methods such as Dynabench~\citep{kiela2021dynabench} and Adaptive Testing~\citep{ribeiro-lundberg-2022-adaptive} iteratively harden test sets. Structured approaches build task hierarchies (DARG~\citep{zhang2024dargdynamicevaluationlarge}, EvalTree~\citep{zeng2025evaltreeprofilinglanguagemodel}, TaskBench~\citep{shen2024taskbenchbenchmarkinglargelanguage}), while autonomous systems such as AutoBencher~\citep{li2025autobencherdeclarativebenchmarkconstruction} and Automated Capability Discovery~\citep{lu2025automated} aim for fully generative benchmarks. Other approaches optimize for particular objectives such as difficulty~\citep{li2024activeevaluationacquisitionefficient}, ethical reasoning~\citep{jiang2025raisingbarinvestigatingvalues,brown2025adaptivelyevaluatingmodelstask}, or adversarial robustness (HarmBench~\citep{mazeika2024harmbenchstandardizedevaluationframework}). These methods reveal important gaps overlooked by static resources but often remain constrained by predefined evaluation goals or reliance on existing benchmarks.

\textbf{Efficiency and Benchmark Optimization.} Benchmark generation is costly, and recent work explores active learning to target the most informative samples. Hassan et al.~\citep{hassan2024activelearningrobustrepresentative} use clustering to expose rare, safety-critical cases, while Li et al.~\citep{li2024activeevaluationacquisitionefficient} introduce RL-based subset selection for efficient evaluation. Despite these advances, many approaches remain tied to fixed datasets or optimize for narrow objectives, leaving open the broader challenge of discovering new capabilities and efficiently approximating performance across large and evolving skill spaces.

\section{Conclusion}
We introduced ACE, a framework for scalable, structured, and efficient evaluation of foundation models. ACE leverages frontier models to construct semantically meaningful capability hierarchies and associated evaluation tasks for a target domain. It further employs active learning in a latent semantic space to efficiently estimate a model's capability function and uncover strengths and weaknesses with minimal evaluation cost.

A limitation of the ACE framework is its reliance on frontier (scientist) models to generate, verify, and score tasks. While practical and scalable, this design raises valid questions about model hallucination, biases, and data contamination; however, employing several scientist models mitigate such risks to some extent. Designing a multi-agent framework where agents debate and critic each other's work  could reduce some of these risks \citep{du2023improving,liang2023encouraging}. In addition, to estimate the true label (e.g., correctness of a generated solution), we can adopt statistical models such as the Dawid–Skene model \citep{dawid1979maximum} or frameworks based on Item Response Theory \citep{baker2001basics}, both of which are designed to aggregate noisy or uncertain judgments from multiple models. Adopting these techniques in the context of LLM-based evaluation is a promising direction for future work.

We believe ACE is a step toward scalable and adaptive evaluation of foundation models. As these models are increasingly deployed in high-stakes domains, the demand for fine-grained and cost-effective evaluation grows. By integrating frontier models with active learning, ACE lays the groundwork for rigorous and reliable evaluation.

\section{Ethics Statement}
ACE reduces human labor and improves scalability in capability evaluation, but it also relies on frontier “scientist models” to generate, verify, and score tasks. This design introduces risks of hallucination, bias in generated content, and potential data contamination from pretraining corpora. We mitigate these risks through multi-pass verification, targeted human inspection, and transparent reporting of limitations. Future extensions of ACE could incorporate multi-agent debate mechanisms or statistical aggregation methods (e.g., Dawid–Skene, Item Response Theory) to further improve robustness.

Our experiments are restricted to mathematics, where problems and solutions can be deterministically verified, minimizing risks of direct harm. While the framework could be extended to sensitive domains such as healthcare or law, such applications should proceed only with domain-specific oversight, ethical safeguards, and regulatory compliance (e.g., IRB approval, privacy protections).

By providing fine-grained, cost-efficient, and extensible evaluation, ACE can improve transparency and reliability in assessing foundation models, uncovering strengths and weaknesses that aggregate metrics overlook. However, uncritical adoption carries risks: benchmarks generated by ACE may inherit biases or errors from underlying models, and use in socially sensitive contexts without safeguards could exacerbate inequities. We emphasize that ACE should be applied responsibly, with human oversight and alignment to community standards on fairness, accountability, and transparency.

\section{Reproducibility Statement}
We have taken several steps to ensure the reproducibility of our work. The ACE framework source code, along with experimental configurations, training/test splits, acquisition functions, and analysis scripts, is open-sourced in our GitHub repository (\url{https://anonymous.4open.science/r/ace-7EAF}). A complete description of generated capabilities, tasks, evaluation procedures, and prompts is provided in the appendix and supplementary materials. In addition, we release JSON files containing all capabilities, areas, and scores produced by the scientist models, along with subject model scores and predictions from active learning. These resources make it possible for researchers to reproduce our benchmark construction end-to-end or to reuse any component of the framework independently.

\bibliography{main}

\begin{thebibliography}{46}
\providecommand{\natexlab}[1]{#1}
\providecommand{\url}[1]{\texttt{#1}}
\expandafter\ifx\csname urlstyle\endcsname\relax
  \providecommand{\doi}[1]{doi: #1}\else
  \providecommand{\doi}{doi: \begingroup \urlstyle{rm}\Url}\fi

\bibitem[{AI Security Institute, UK}(2024)]{inspect}
{AI Security Institute, UK}.
\newblock {Inspect AI: Framework for Large Language Model Evaluations}.
\newblock \url{https://github.com/UKGovernmentBEIS/inspect_ai}, 2024.
\newblock Accessed: 2024-05.

\bibitem[Baker(2001)]{baker2001basics}
Frank~B Baker.
\newblock \emph{The basics of item response theory}.
\newblock ERIC, 2001.

\bibitem[Brown et~al.(2025)Brown, Balehannina, Jin, Havaldar, Hassani, and Wong]{brown2025adaptivelyevaluatingmodelstask}
Davis Brown, Prithvi Balehannina, Helen Jin, Shreya Havaldar, Hamed Hassani, and Eric Wong.
\newblock Adaptively evaluating models with task elicitation, 2025.
\newblock URL \url{https://arxiv.org/abs/2503.01986}.

\bibitem[Chabanet et~al.(2021)Chabanet, El-Haouzi, and Thomas]{chabanet2021coupling}
Sylvain Chabanet, Hind~Bril El-Haouzi, and Philippe Thomas.
\newblock Coupling digital simulation and machine learning metamodel through an active learning approach in industry 4.0 context.
\newblock \emph{Computers in Industry}, 133:\penalty0 103529, 2021.

\bibitem[Chen et~al.(2021)Chen, Tworek, Jun, Yuan, Pinto, Kaplan, Edwards, Burda, Joseph, Brockman, et~al.]{chen2021evaluating}
Mark Chen, Jerry Tworek, Heewoo Jun, Qiming Yuan, Henrique Ponde De~Oliveira Pinto, Jared Kaplan, Harri Edwards, Yuri Burda, Nicholas Joseph, Greg Brockman, et~al.
\newblock Evaluating large language models trained on code.
\newblock \emph{arXiv preprint arXiv:2107.03374}, 2021.

\bibitem[Clark et~al.(2018)Clark, Cowhey, Etzioni, Khot, Sabharwal, Schoenick, and Tafjord]{clark2018think}
Peter Clark, Isaac Cowhey, Oren Etzioni, Tushar Khot, Ashish Sabharwal, Carissa Schoenick, and Oyvind Tafjord.
\newblock Think you have solved question answering? try arc, the ai2 reasoning challenge.
\newblock \emph{arXiv preprint arXiv:1803.05457}, 2018.

\bibitem[Cobbe et~al.(2021)Cobbe, Kosaraju, Bavarian, Chen, Jun, Kaiser, Plappert, Tworek, Hilton, Nakano, et~al.]{cobbe2021training}
Karl Cobbe, Vineet Kosaraju, Mohammad Bavarian, Mark Chen, Heewoo Jun, Lukasz Kaiser, Matthias Plappert, Jerry Tworek, Jacob Hilton, Reiichiro Nakano, et~al.
\newblock Training verifiers to solve math word problems, 2021.
\newblock \emph{URL https://arxiv. org/abs/2110.14168}, 9, 2021.

\bibitem[Cohn(1996)]{cohn1996neural}
David~A Cohn.
\newblock Neural network exploration using optimal experiment design.
\newblock \emph{Neural networks}, 9\penalty0 (6):\penalty0 1071--1083, 1996.

\bibitem[Dawid \& Skene(1979)Dawid and Skene]{dawid1979maximum}
Alexander~Philip Dawid and Allan~M Skene.
\newblock Maximum likelihood estimation of observer error-rates using the em algorithm.
\newblock \emph{Journal of the Royal Statistical Society: Series C (Applied Statistics)}, 28\penalty0 (1):\penalty0 20--28, 1979.

\bibitem[Deng et~al.(2024)Deng, Zhao, Tang, Gerstein, and Cohan]{deng2024investigatingdatacontaminationmodern}
Chunyuan Deng, Yilun Zhao, Xiangru Tang, Mark Gerstein, and Arman Cohan.
\newblock Investigating data contamination in modern benchmarks for large language models, 2024.
\newblock URL \url{https://arxiv.org/abs/2311.09783}.

\bibitem[Du et~al.(2023)Du, Li, Torralba, Tenenbaum, and Mordatch]{du2023improving}
Yilun Du, Shuang Li, Antonio Torralba, Joshua~B Tenenbaum, and Igor Mordatch.
\newblock Improving factuality and reasoning in language models through multiagent debate, 2023.
\newblock \emph{URL https://arxiv. org/abs/2305.14325}, 3, 2023.

\bibitem[Dua et~al.(2019)Dua, Wang, Dasigi, Stanovsky, Singh, and Gardner]{dua2019drop}
Dheeru Dua, Yizhong Wang, Pradeep Dasigi, Gabriel Stanovsky, Sameer Singh, and Matt Gardner.
\newblock Drop: A reading comprehension benchmark requiring discrete reasoning over paragraphs.
\newblock \emph{arXiv preprint arXiv:1903.00161}, 2019.

\bibitem[Fu(2022)]{fu2022active}
Siwei Fu.
\newblock Active learning for solving expensive optimization problems, 2022.

\bibitem[Gorissen et~al.(2009)Gorissen, Crombecq, Couckuyt, and Dhaene]{gorissen2009automatic}
Dirk Gorissen, Karel Crombecq, Ivo Couckuyt, and Tom Dhaene.
\newblock Automatic approximation of expensive functions with active learning.
\newblock \emph{Foundations of Computational, Intelligence Volume 1: Learning and Approximation}, pp.\  35--62, 2009.

\bibitem[Hassan et~al.(2024)Hassan, Sicilia, and Alikhani]{hassan2024activelearningrobustrepresentative}
Sabit Hassan, Anthony Sicilia, and Malihe Alikhani.
\newblock Active learning for robust and representative llm generation in safety-critical scenarios, 2024.
\newblock URL \url{https://arxiv.org/abs/2410.11114}.

\bibitem[Hendrycks et~al.(2020)Hendrycks, Burns, Basart, Zou, Mazeika, Song, and Steinhardt]{hendrycks2020measuring}
Dan Hendrycks, Collin Burns, Steven Basart, Andy Zou, Mantas Mazeika, Dawn Song, and Jacob Steinhardt.
\newblock Measuring massive multitask language understanding.
\newblock \emph{arXiv preprint arXiv:2009.03300}, 2020.

\bibitem[Hendrycks et~al.(2021{\natexlab{a}})Hendrycks, Burns, Basart, Zou, Mazeika, Song, and Steinhardt]{hendrycks2021measuringmassivemultitasklanguage}
Dan Hendrycks, Collin Burns, Steven Basart, Andy Zou, Mantas Mazeika, Dawn Song, and Jacob Steinhardt.
\newblock Measuring massive multitask language understanding, 2021{\natexlab{a}}.
\newblock URL \url{https://arxiv.org/abs/2009.03300}.

\bibitem[Hendrycks et~al.(2021{\natexlab{b}})Hendrycks, Burns, Kadavath, Arora, Basart, Tang, Song, and Steinhardt]{hendrycks2021measuring}
Dan Hendrycks, Collin Burns, Saurav Kadavath, Akul Arora, Steven Basart, Eric Tang, Dawn Song, and Jacob Steinhardt.
\newblock Measuring mathematical problem solving with the math dataset.
\newblock \emph{NeurIPS}, 2021{\natexlab{b}}.

\bibitem[Ili{\'c} \& Gignac(2024)Ili{\'c} and Gignac]{ilic2024evidence}
David Ili{\'c} and Gilles~E Gignac.
\newblock Evidence of interrelated cognitive-like capabilities in large language models: Indications of artificial general intelligence or achievement?
\newblock \emph{Intelligence}, 106:\penalty0 101858, 2024.

\bibitem[Jiang et~al.(2025)Jiang, Yi, Wei, Xiao, Wang, and Xie]{jiang2025raisingbarinvestigatingvalues}
Han Jiang, Xiaoyuan Yi, Zhihua Wei, Ziang Xiao, Shu Wang, and Xing Xie.
\newblock Raising the bar: Investigating the values of large language models via generative evolving testing, 2025.
\newblock URL \url{https://arxiv.org/abs/2406.14230}.

\bibitem[Kiela et~al.(2021)Kiela, Bartolo, Nie, Kaushik, Geiger, Wu, Vidgen, Prasad, Singh, Ringshia, et~al.]{kiela2021dynabench}
Douwe Kiela, Max Bartolo, Yixin Nie, Divyansh Kaushik, Atticus Geiger, Zhengxuan Wu, Bertie Vidgen, Grusha Prasad, Amanpreet Singh, Pratik Ringshia, et~al.
\newblock Dynabench: Rethinking benchmarking in nlp.
\newblock \emph{arXiv preprint arXiv:2104.14337}, 2021.

\bibitem[Li et~al.(2025)Li, Kaiyom, Liu, Mai, Liang, and Hashimoto]{li2025autobencherdeclarativebenchmarkconstruction}
Xiang~Lisa Li, Farzaan Kaiyom, Evan~Zheran Liu, Yifan Mai, Percy Liang, and Tatsunori Hashimoto.
\newblock Autobencher: Towards declarative benchmark construction, 2025.
\newblock URL \url{https://arxiv.org/abs/2407.08351}.

\bibitem[Li et~al.(2024)Li, Ma, Ballesteros, Benajiba, and Horwood]{li2024activeevaluationacquisitionefficient}
Yang Li, Jie Ma, Miguel Ballesteros, Yassine Benajiba, and Graham Horwood.
\newblock Active evaluation acquisition for efficient llm benchmarking, 2024.
\newblock URL \url{https://arxiv.org/abs/2410.05952}.

\bibitem[Liang et~al.(2022)Liang, Bommasani, Lee, Tsipras, Soylu, Yasunaga, Zhang, Narayanan, Wu, Kumar, et~al.]{liang2022holistic}
Percy Liang, Rishi Bommasani, Tony Lee, Dimitris Tsipras, Dilara Soylu, Michihiro Yasunaga, Yian Zhang, Deepak Narayanan, Yuhuai Wu, Ananya Kumar, et~al.
\newblock Holistic evaluation of language models.
\newblock \emph{arXiv preprint arXiv:2211.09110}, 2022.

\bibitem[Liang et~al.(2023)Liang, He, Jiao, Wang, Wang, Wang, Yang, Shi, and Tu]{liang2023encouraging}
Tian Liang, Zhiwei He, Wenxiang Jiao, Xing Wang, Yan Wang, Rui Wang, Yujiu Yang, Shuming Shi, and Zhaopeng Tu.
\newblock Encouraging divergent thinking in large language models through multi-agent debate.
\newblock \emph{arXiv preprint arXiv:2305.19118}, 2023.

\bibitem[Lin et~al.(2022)Lin, Hilton, and Evans]{lin2022truthfulqameasuringmodelsmimic}
Stephanie Lin, Jacob Hilton, and Owain Evans.
\newblock Truthfulqa: Measuring how models mimic human falsehoods, 2022.
\newblock URL \url{https://arxiv.org/abs/2109.07958}.

\bibitem[Lu et~al.(2025)Lu, Hu, and Clune]{lu2025automated}
Cong Lu, Shengran Hu, and Jeff Clune.
\newblock Automated capability discovery via model self-exploration.
\newblock \emph{arXiv preprint arXiv:2502.07577}, 2025.

\bibitem[MacKay(1992)]{mackay1992information}
David~JC MacKay.
\newblock Information-based objective functions for active data selection.
\newblock \emph{Neural computation}, 4\penalty0 (4):\penalty0 590--604, 1992.

\bibitem[Malkomes(2019)]{malkomes2019automating}
Gustavo Malkomes.
\newblock Automating active learning for gaussian processes.
\newblock 2019.

\bibitem[Mazeika et~al.(2024)Mazeika, Phan, Yin, Zou, Wang, Mu, Sakhaee, Li, Basart, Li, Forsyth, and Hendrycks]{mazeika2024harmbenchstandardizedevaluationframework}
Mantas Mazeika, Long Phan, Xuwang Yin, Andy Zou, Zifan Wang, Norman Mu, Elham Sakhaee, Nathaniel Li, Steven Basart, Bo~Li, David Forsyth, and Dan Hendrycks.
\newblock Harmbench: A standardized evaluation framework for automated red teaming and robust refusal, 2024.
\newblock URL \url{https://arxiv.org/abs/2402.04249}.

\bibitem[Phan et~al.(2025)Phan, Gatti, Han, Li, Hu, Zhang, Zhang, Shaaban, Ling, Shi, et~al.]{phan2025humanity}
Long Phan, Alice Gatti, Ziwen Han, Nathaniel Li, Josephina Hu, Hugh Zhang, Chen Bo~Calvin Zhang, Mohamed Shaaban, John Ling, Sean Shi, et~al.
\newblock Humanity's last exam.
\newblock \emph{arXiv preprint arXiv:2501.14249}, 2025.

\bibitem[Prabhu et~al.(2024)Prabhu, Udandarao, Torr, Bethge, Bibi, and Albanie]{prabhu2024efficient}
Ameya Prabhu, Vishaal Udandarao, Philip Torr, Matthias Bethge, Adel Bibi, and Samuel Albanie.
\newblock Efficient lifelong model evaluation in an era of rapid progress.
\newblock \emph{Advances in Neural Information Processing Systems}, 37:\penalty0 74089--74121, 2024.

\bibitem[Rasmussen \& Williams(2006)Rasmussen and Williams]{rasmussen2006gaussian}
Carl~Edward Rasmussen and Christopher~KI Williams.
\newblock \emph{Gaussian processes for machine learning}, volume~2.
\newblock MIT press Cambridge, MA, 2006.

\bibitem[Ribeiro \& Lundberg(2022)Ribeiro and Lundberg]{ribeiro-lundberg-2022-adaptive}
Marco~Tulio Ribeiro and Scott Lundberg.
\newblock Adaptive testing and debugging of {NLP} models.
\newblock In Smaranda Muresan, Preslav Nakov, and Aline Villavicencio (eds.), \emph{Proceedings of the 60th Annual Meeting of the Association for Computational Linguistics (Volume 1: Long Papers)}, pp.\  3253--3267, Dublin, Ireland, May 2022. Association for Computational Linguistics.
\newblock \doi{10.18653/v1/2022.acl-long.230}.
\newblock URL \url{https://aclanthology.org/2022.acl-long.230/}.

\bibitem[Riis et~al.(2021)Riis, Antunes, Gurtner, Pereira, Delgado, and Azevedo]{riis2021active}
Christoffer Riis, Francisco Antunes, G{\'e}rald Gurtner, Francisco~Camara Pereira, Luis Delgado, and Carlos M~Lima Azevedo.
\newblock Active learning metamodels for atm simulation modeling.
\newblock In \emph{11th SESAR Innovation Days}, 2021.

\bibitem[Seeger(2002)]{seeger2002pac}
Matthias Seeger.
\newblock Pac-bayesian generalisation error bounds for gaussian process classification.
\newblock \emph{Journal of machine learning research}, 3\penalty0 (Oct):\penalty0 233--269, 2002.

\bibitem[Seo et~al.(2000)Seo, Wallat, Graepel, and Obermayer]{seo2000gaussian}
Sambu Seo, Marko Wallat, Thore Graepel, and Klaus Obermayer.
\newblock Gaussian process regression: Active data selection and test point rejection.
\newblock In \emph{Mustererkennung 2000}, pp.\  27--34. Springer, 2000.

\bibitem[Shen et~al.(2024)Shen, Song, Tan, Zhang, Ren, Yuan, Lu, Li, and Zhuang]{shen2024taskbenchbenchmarkinglargelanguage}
Yongliang Shen, Kaitao Song, Xu~Tan, Wenqi Zhang, Kan Ren, Siyu Yuan, Weiming Lu, Dongsheng Li, and Yueting Zhuang.
\newblock Taskbench: Benchmarking large language models for task automation, 2024.
\newblock URL \url{https://arxiv.org/abs/2311.18760}.

\bibitem[Siska et~al.(2024)Siska, Marazopoulou, Ailem, and Bono]{siska2024examining}
Charlotte Siska, Katerina Marazopoulou, Melissa Ailem, and James Bono.
\newblock Examining the robustness of llm evaluation to the distributional assumptions of benchmarks.
\newblock In \emph{Proceedings of the 62nd Annual Meeting of the Association for Computational Linguistics (Volume 1: Long Papers)}, pp.\  10406--10421, 2024.

\bibitem[Srivastava et~al.(2022)Srivastava, Rastogi, Rao, Shoeb, Abid, Fisch, Brown, Santoro, Gupta, Garriga-Alonso, et~al.]{srivastava2022beyond}
Aarohi Srivastava, Abhinav Rastogi, Abhishek Rao, Abu Awal~Md Shoeb, Abubakar Abid, Adam Fisch, Adam~R Brown, Adam Santoro, Aditya Gupta, Adri{\`a} Garriga-Alonso, et~al.
\newblock Beyond the imitation game: Quantifying and extrapolating the capabilities of language models.
\newblock \emph{arXiv preprint arXiv:2206.04615}, 2022.

\bibitem[Van~der Maaten \& Hinton(2008)Van~der Maaten and Hinton]{van2008visualizing}
Laurens Van~der Maaten and Geoffrey Hinton.
\newblock Visualizing data using t-sne.
\newblock \emph{Journal of machine learning research}, 9\penalty0 (11), 2008.

\bibitem[Wang et~al.(2024)Wang, Wu, Ma, and Liu]{wang2024fac}
Xiaoqiang Wang, Lingfei Wu, Tengfei Ma, and Bang Liu.
\newblock Fac$^2$e: Better understanding large language model capabilities by dissociating language and cognition.
\newblock \emph{arXiv preprint arXiv:2403.00126}, 2024.

\bibitem[Zellers et~al.(2019)Zellers, Holtzman, Bisk, Farhadi, and Choi]{zellers2019hellaswag}
Rowan Zellers, Ari Holtzman, Yonatan Bisk, Ali Farhadi, and Yejin Choi.
\newblock Hellaswag: Can a machine really finish your sentence?
\newblock \emph{arXiv preprint arXiv:1905.07830}, 2019.

\bibitem[Zeng et~al.(2025)Zeng, Wang, Hajishirzi, and Koh]{zeng2025evaltreeprofilinglanguagemodel}
Zhiyuan Zeng, Yizhong Wang, Hannaneh Hajishirzi, and Pang~Wei Koh.
\newblock Evaltree: Profiling language model weaknesses via hierarchical capability trees, 2025.
\newblock URL \url{https://arxiv.org/abs/2503.08893}.

\bibitem[Zhang et~al.(2024{\natexlab{a}})Zhang, Chen, and Yang]{zhang2024darg}
Zhehao Zhang, Jiaao Chen, and Diyi Yang.
\newblock Darg: Dynamic evaluation of large language models via adaptive reasoning graph.
\newblock \emph{Advances in Neural Information Processing Systems}, 37:\penalty0 135904--135942, 2024{\natexlab{a}}.

\bibitem[Zhang et~al.(2024{\natexlab{b}})Zhang, Chen, and Yang]{zhang2024dargdynamicevaluationlarge}
Zhehao Zhang, Jiaao Chen, and Diyi Yang.
\newblock Darg: Dynamic evaluation of large language models via adaptive reasoning graph, 2024{\natexlab{b}}.
\newblock URL \url{https://arxiv.org/abs/2406.17271}.

\end{thebibliography}
\bibliographystyle{iclr2026_conference}

\newpage
\appendix
\section*{Appendix}

\section{Active Learning for Capability Function Approximation}
\label{app:alg}
\begin{algorithm}[htb]
\caption{Active Learning for Capability Function Approximation}
\label{alg:ace}
\SetAlgoLined
\KwIn{ \\
\hspace{5mm} Initial capability set $\mathcal{C} = \{c_i\}_{i=1}^N$ generated by the scientist models \\
\hspace{5mm} Pretrained encoder $E: \mathcal{C} \rightarrow \mathbb{R}^d$ \\
\hspace{5mm} Dimensionality reduction method $\varphi$ (e.g., PCA, t-SNE) \\
\hspace{5mm} Evaluation module $\texttt{Evaluate()}$ to score a capability \\
\hspace{5mm} Active learning acquisition function $\alpha(\cdot)$ \\
\hspace{5mm} Target latent dimension $d' \ll d$
}

\BlankLine

\textbf{Initialization:} \\
1. Encode all capabilities: $\mathbf{Z} = \{E(c_i) | c_i \in \mathcal{C}\}$ \\
2. Reduce dimensionality: $\mathbf{Z}' = \varphi(\mathbf{Z}) \in \mathbb{R}^{N \times d'}$ \\
3. Initialize training set $\mathcal{D}$ by randomly selecting a small number of capabilities (e.g., 2) from $\mathcal{C}$ and scoring them using \texttt{Evaluate()} \\

\BlankLine

\texttt{// Active learning} \\
\While{stopping conditions not met}{
  1. Fit GP model, $f$, on current $\mathcal{D}$ (non-parametric) \\
  2. Compute acquisition scores: $\forall \mathbf{z}'_i \in \mathbf{Z}' \setminus \mathcal{D}, \alpha_i \leftarrow \alpha(\mathbf{z}'_i; f)$ \\
  3. Select the best candidate: $j \leftarrow \argmax_i \alpha_i$ \\
  4. Obtain capability score: $s_j \leftarrow \texttt{Evaluate}(c_j)$ \\
  5. Update training set: $\mathcal{D} \leftarrow \mathcal{D} \cup \{(\mathbf{z}'_j, s_j)\}$ \\
}

\Return $\mathcal{D}$
\end{algorithm}

\section{Active Learning with Gaussian Processes}
\label{app:active_learning_gp}

A Gaussian process (GP) is a collection of random variables, any finite number of which have a joint Gaussian distribution \citep{rasmussen2006gaussian}. It is fully specified by a mean function $m(\mathbf{x}) = \mathbb{E}[f(\mathbf{x})]$ and a covariance (kernel) function $k(\mathbf{x}, \mathbf{x}') = \mathbb{E}[(f(\mathbf{x}) - m(\mathbf{x}))(f(\mathbf{x}') - m(\mathbf{x}'))]$:
\beqann
    f(\mathbf{x}) \sim \mathcal{GP}(m(\mathbf{x}), k(\mathbf{x}, \mathbf{x}'))
\eeqann

Consider a regression task with training data $\mathcal{D} = \{(\mathbf{x}_i, y_i)\}_{i=1}^N$ where $y_i = f(\mathbf{x}_i) + \epsilon_i$ with $\epsilon_i \sim \mathcal{N}(0, \sigma_n^2)$. For a test input $\mathbf{x}_*$, the predictive distribution is Gaussian:
\beqann
    p(f_*|\mathbf{x}_*, \mathcal{D}) = \mathcal{N}(\mathbb{E}[f_*], \mathbb{V}[f_*]),
\eeqann
with predictive mean and variance:
\beqa
    \mathbb{E}[f_*] &=& \mathbf{k}_*\transpose(\mathbf{K} + \sigma_n^2\mathbf{I})^{-1}\mathbf{y} \\
    \mathbb{V}[f_*] &=& k(\mathbf{x}_*, \mathbf{x}_*) - \mathbf{k}_*\transpose(\mathbf{K} + \sigma_n^2\mathbf{I})^{-1}\mathbf{k}_*,
\label{eq:var}
\eeqa
in which $\mathbf{K}$ is the kernel matrix with $K_{ij} = k(\mathbf{x}_i, \mathbf{x}_j)$, $\mathbf{y} = \{y_1,\ldots,y_N\}$, and $\mathbf{k}_* = [k(\mathbf{x}_1, \mathbf{x}_*), ..., k(\mathbf{x}_N, \mathbf{x}_*)]\transpose$.

The function-space view interprets the GP as defining a distribution over functions, where the kernel function encodes prior assumptions such as smoothness. A common choice is the squared exponential kernel:
\beqann
    k(\mathbf{x}, \mathbf{x}') = \sigma_f^2 \exp\left(-\frac{||\mathbf{x} - \mathbf{x}'||^2}{2l^2}\right).
\eeqann

GPs naturally lend themselves to active learning due to the availability of posterior mean and variance estimates.
In particular two well-known approaches leverage GP posterior variance for active learning. \citet{mackay1992information} aims at maximizing the expected information gain by selecting the data where the model has maximum variance. This is performed by selecting points that maximize the posterior variance:
\beqa
    \mathbf{x}^* = \argmax_{\mathbf{x} \in \mathcal{U}} \mathbb{V}[f(\mathbf{x})],
\label{eq:mackay}
\eeqa
where $\mathcal{U}$ is the pool of unlabeled candidates. This is equivalent to maximizing the reduction in entropy $H$ of the GP posterior:
\beqann
    \mathbf{x}^* = \argmax_{\mathbf{x} \in \mathcal{U}} H[p(f|\mathcal{D})] - \mathbb{E}_{y|\mathbf{x}}[H[p(f|\mathcal{D} \cup (\mathbf{x}, y))]].
\eeqann
It is possible to perform optimization of Eq.~\ref{eq:var} with respect to $\mathbf{x}^*$ using, e.g., gradient ascent~\citep{seo2000gaussian}.

The second method is motivated by minimizing the generalization error in terms of mean squared error (MSE). Using the bias-variance decomposition of MSE and making some assumptions with respect to the magnitude of bias, it can be shown that minimizing MSE can be approximated by choosing the candidate point that reduces the expected predictive variance over the entire input space \citep{cohn1996neural}:
\beqa
    \mathbf{x}^* = \argmin_{\mathbf{x} \in \mathcal{U}} \mathbb{E}_{y|\mathbf{x}}\left[\int \mathbb{V}[f(\mathbf{x}')|\mathcal{D} \cup (\mathbf{x}, y)] d\mathbf{x}'\right]
\label{eq:cohn}
\eeqa
In practice the integration in Eq.~\ref{eq:cohn} can be approximated by Monte Carlo or by calculating the variance over a holdout set.

For GPs, both approaches can be approximated efficiently as the posterior covariance matrix can be updated incrementally using rank-1 updates \citep{seeger2002pac}. The active learning process iteratively fits the GP to current labeled data, $\mathcal{L}$, computes the acquisition score (Eq.~\ref{eq:mackay} or~\ref{eq:cohn}) for all $\mathbf{x} \in \mathcal{U}$, selects $\mathbf{x}^*$ that maximizes the acquisition score, queries for $y^*$ at $\mathbf{x}^*$, and updates the labeled and candidate sets, $\mathcal{L} \leftarrow \mathcal{L} \cup \{(\mathbf{x}^*, y^*)\}$, $\mathcal{U} \leftarrow \mathcal{U} \setminus \{\mathbf{x}^*\}$.

\section{Capability Details}
In this section, we provide details on the generated capabilities, their embeddings used in our method, and LLM scores evaluated on each capability.
\subsection{Capability Embeddings}
\begin{figure}[tb]
    \centering
    \includegraphics[width=0.36\textwidth]{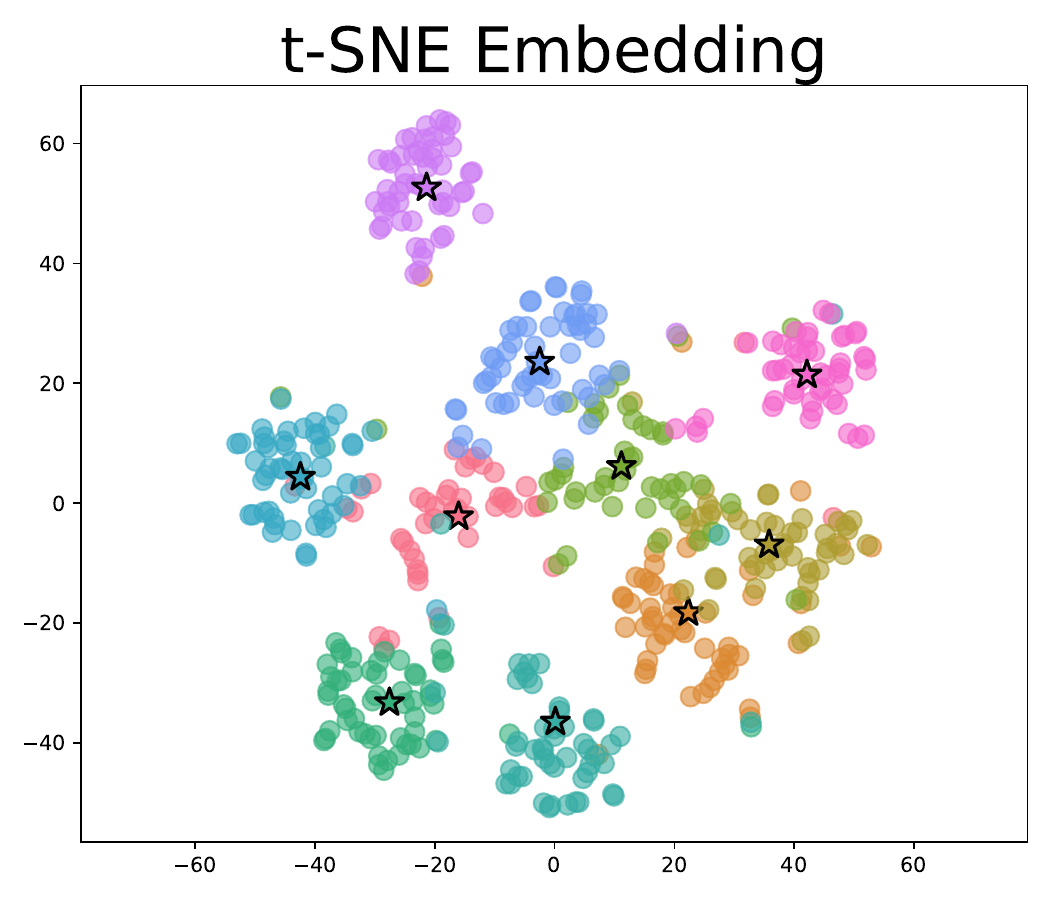}
    \includegraphics[width=0.36\textwidth]{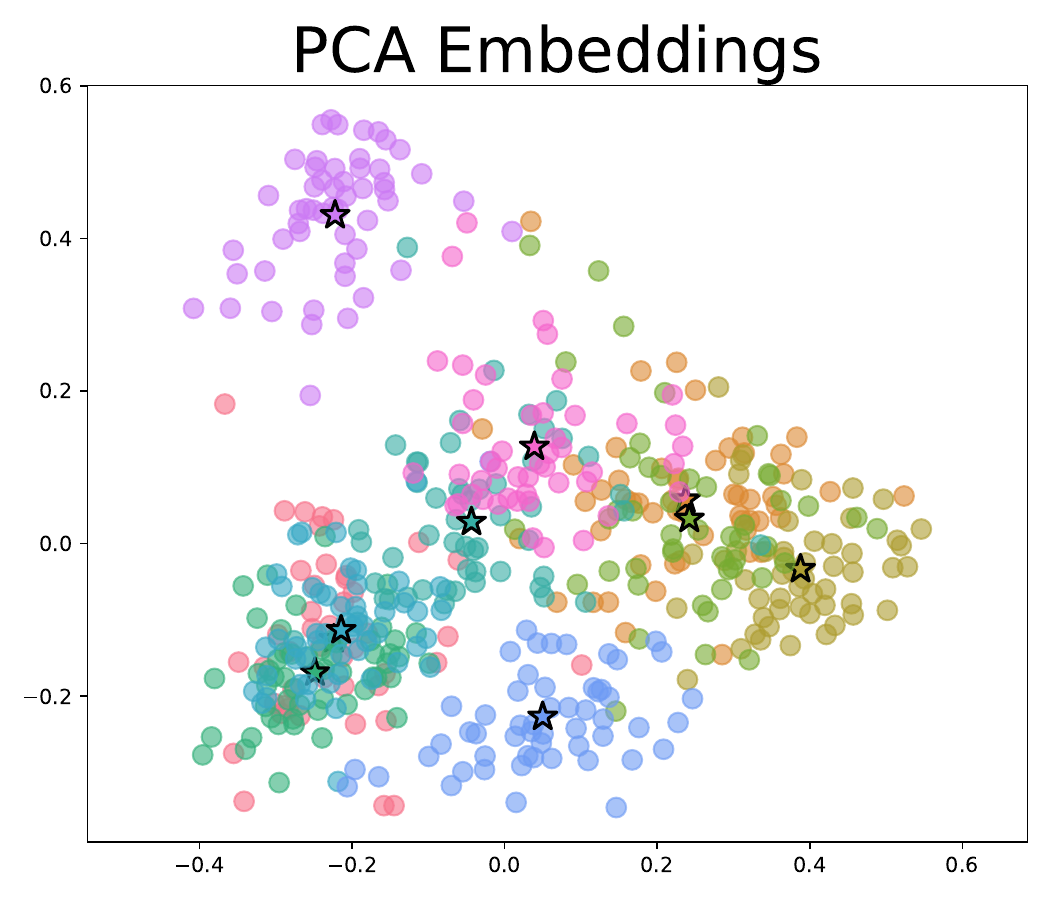}
    \raisebox{0.28cm}{\includegraphics[width=0.26\textwidth]{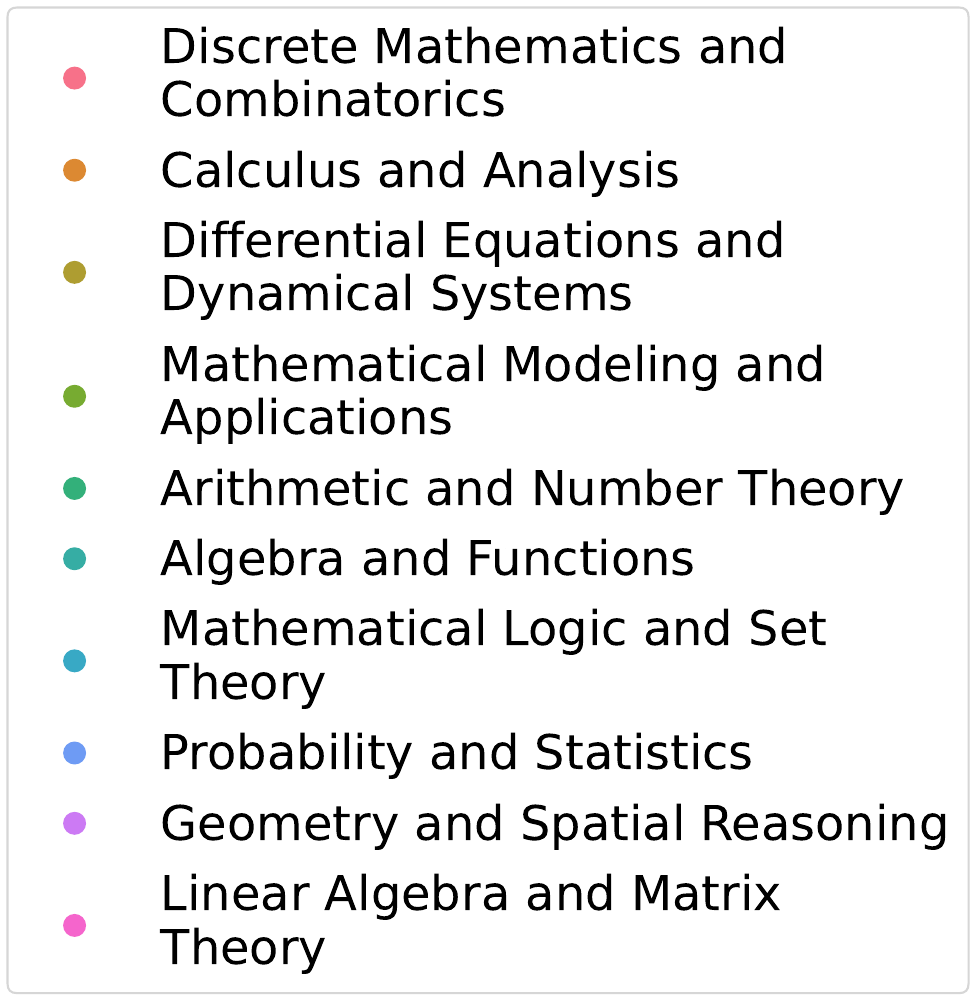}}
    \caption{Two-dimensional representation of Mathematics capabilities using t-SNE (left) and PCA (right). Each point corresponds to a capability, and colors indicate high-level areas. Stars indicate the mean of capability representations for each area.}
    \label{fig:capability_distribution}
\end{figure}

Figure~\ref{fig:capability_distribution} shows the distribution of capabilities in a latent 2D space for t-SNE (left) and PCA (right) dimensionality reduction techniques.

\subsection{Capability Similarity Heatmap}
\label{app:capability_embedding_heatmap}
\begin{figure}
    \centering
    \includegraphics[width=0.99\linewidth]{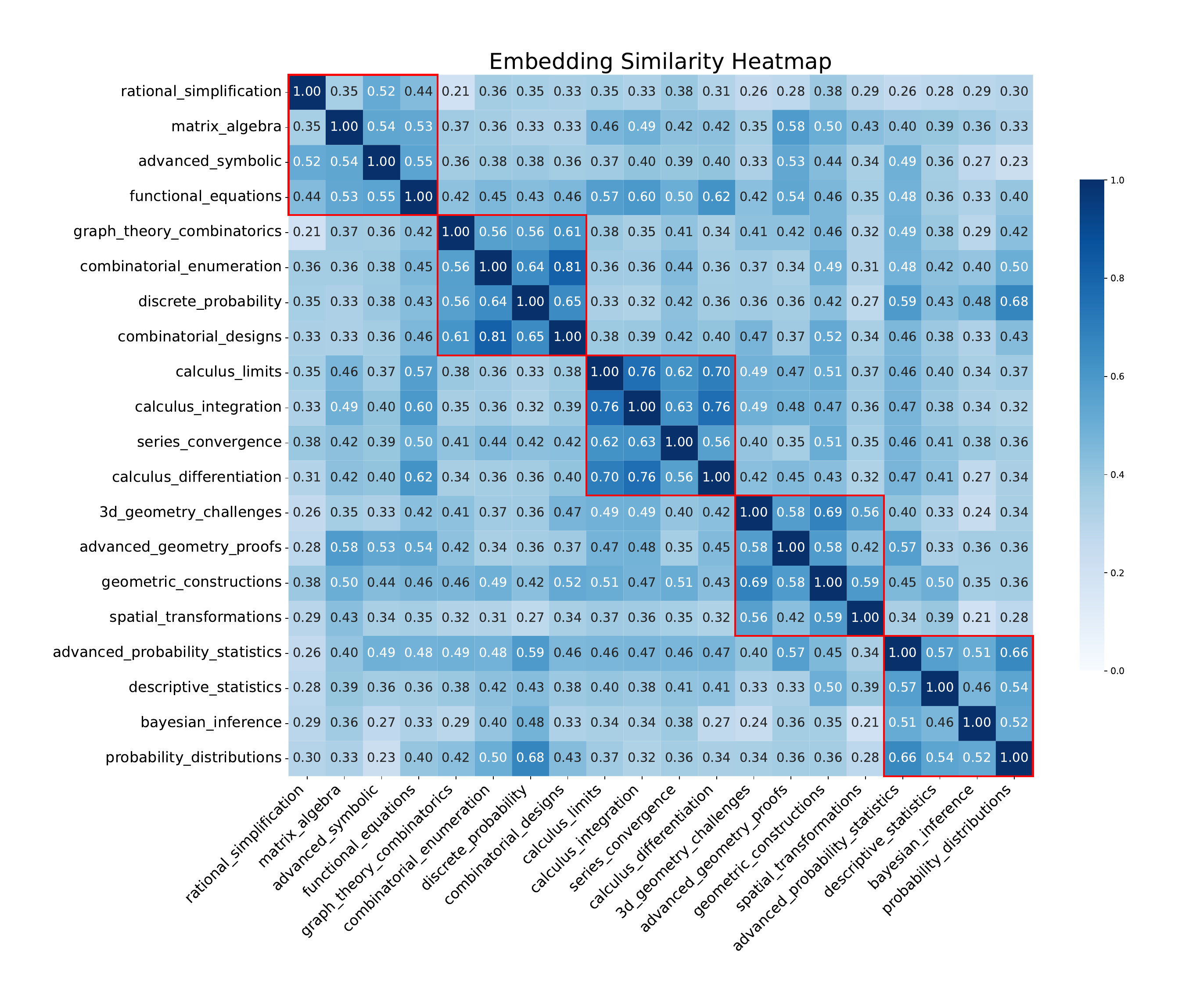}
    \caption{The heatmap illustrating the cosine similarity matrix of capability embeddings. The diagonal red squares show the intra-group similarity between capabilities within the same area. }
    \label{fig:embedding_hetamap}
\end{figure}

As expected, capability embeddings generated from capability names, areas, and descriptions carry the semantic similarity of the capabilities. Therefore, as the heatmap in Figure \ref{fig:embedding_hetamap} shows, capability embeddings within the same area have higher cosine similarity compared to the capabilities in other areas. 

\section{Manual Inspection of Tasks}\label{sec:manual_inspection}

To evaluate the quality of the task generation pipeline and the reliability of the automated verification step, we conducted a manual inspection of a subset of tasks. Specifically, we randomly selected 12 capabilities across three mathematical areas.
For each capability, we sampled 15 tasks, resulting in a total of 180 problem–solution pairs. Each task’s problem, solution, and verification model output were manually reviewed by solving the problem and comparing the correct solution to the automated verification outcome.

The results indicate a high degree of agreement between human and automated verification. Of the 180 tasks, we observed the following confusion matrix: True Positives = 158, False Negatives = 14, False Positives = 1, and True Negatives = 7. This corresponds to a precision of \textbf{99.4\%}, recall of \textbf{91.9\%}, and overall verification accuracy of \textbf{91.7\%}. These results support the conclusion that the automated pipeline for task generation and verification is reliable for evaluating model capabilities at scale. Despite strong performance in task generation, our inspection surfaced a few recurring issues that are important to address in future iterations of the framework:
\begin{enumerate}
\item \textbf{Rounding Errors.}
Infrequent but notable rounding inaccuracies occurred when intermediate numerical results were used in subsequent calculations. These rounding issues sometimes led to small deviations in final answers and highlight the need for improved numerical precision handling.

\item \textbf{Lack of Task Diversity.}  
Many tasks within a capability were structurally or conceptually similar. Increasing task diversity—across difficulty levels and subtopics—would yield a more comprehensive assessment of model performance.

\item \textbf{Inter-Task Dependencies.}  
Since multiple tasks were generated from a single prompt (to minimize repetition), some questions inadvertently referenced earlier tasks. Future prompts should explicitly enforce task independence to avoid this issue.

\item \textbf{Parsing Limitations.}  
Some task-solving instructions required the model to output the final answer after an "ANSWER" keyword. The current parsing logic does not support multi-line answers, which can result in incomplete ground truth extraction and premature task rejection during verification. Improving parsing robustness would reduce unnecessary filtering of valid tasks.
\end{enumerate}

Addressing these issues will further enhance the robustness and reliability of automated task generation and verification.

\section{Fine-Grained Evaluation of Active Learning}
\begin{figure}
    \centering
    \includegraphics[width=0.5\linewidth]{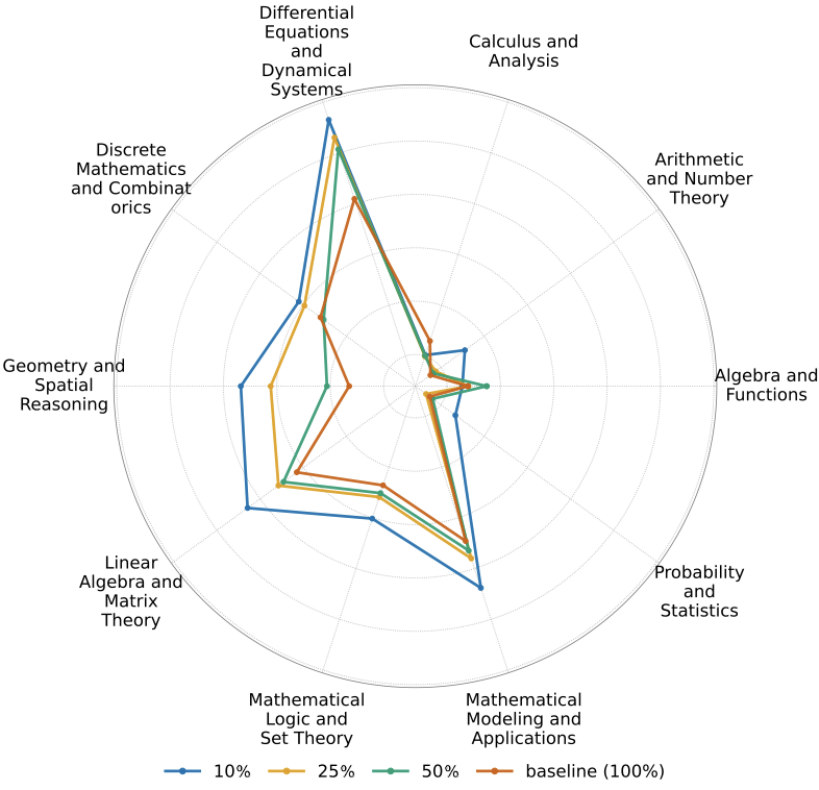}
    \caption{Area level evaluation of the subject model over fractions of training data via active learning.}
    \label{fig:al_spider_chart}
\end{figure}

\section{Prompts}
\label{app:prompts}

\subsection{Capability Area Generation Prompts}\label{sec:area_gen_prompts}

\begin{tcolorbox}[
  colback=promptheaderbg,       
  colframe=promptheaderbg,       
  width=\linewidth,     
  boxrule=0.5pt,        
  boxsep=0pt,
  left=5pt, right=5pt, top=5pt, bottom=5pt,
  arc=4pt, 
  before skip=10pt,
  after skip=0pt,
  enhanced,
  sharp corners=south,
  listing only,         
  listing options={basicstyle=\ttfamily\small}
]
\color{white}\textbf{Capability Area Generation User Prompt}
\end{tcolorbox}
\begin{tcolorbox}[
  colback=promptframebg,       
  colframe=promptheaderbg,       
  width=\linewidth,     
  boxrule=0.5pt,        
  boxsep=0pt,
  left=5pt, right=5pt, top=5pt, bottom=5pt,
  arc=0pt,
  before skip=0pt,
  after skip=10pt,
  enhanced,
  sharp corners,        
  listing only,         
  listing options={basicstyle=\ttfamily\small}
]
You are an expert in designing capabilities to assess the abilities of large language models (LLMs). Identify \texttt{num\_areas} broad and diverse areas for capability generation for the \texttt{domain} domain. Each area should cover \texttt{num\_capabilities\_per\_area} capabilities, which will be generated in the next step. The areas should be relevant to the \texttt{domain} domain, should be high level and should not overlap with each other. \\

Respond precisely in the following format: \\

RESPONSE JSON:

\begin{verbatim}
{
    "area_0": <STR>,
    "area_1": <STR>,
    ...
}
\end{verbatim}
\end{tcolorbox}

\clearpage

\subsection{Capability Generation Prompts}\label{sec:cap_gen_prompts}

\begin{tcolorbox}[
  colback=promptheaderbg,       
  colframe=promptheaderbg,       
  width=\linewidth,     
  boxrule=0.5pt,        
  boxsep=0pt,
  left=5pt, right=5pt, top=5pt, bottom=5pt,
  arc=4pt, 
  before skip=10pt,
  after skip=0pt,
  enhanced,
  sharp corners=south,        
  listing only,         
  listing options={basicstyle=\ttfamily\small}
]
\color{white}\textbf{Capability Generation System Prompt}
\end{tcolorbox}
\begin{tcolorbox}[
  colback=promptframebg,       
  colframe=promptheaderbg,       
  width=\linewidth,     
  boxrule=0.5pt,        
  boxsep=0pt,
  left=5pt, right=5pt, top=5pt, bottom=5pt,
  arc=0pt,
  before skip=0pt,
  after skip=10pt,
  enhanced,
  breakable,
  sharp corners,        
  listing only,         
  listing options={basicstyle=\ttfamily\small}
]
You are an expert in designing capabilities to assess the abilities of large language models (LLMs). Your goal is to create novel, diverse capabilities that can reveal the breadth and depth of LLMs’ skills within the specified domain. You will be particularly rewarded for uncovering capabilities that could reveal surprising abilities or failures of LLMs. Valid capabilities will be added to a capability archive. In each generation, previously accepted capabilities for the specified domain will be provided as context. \\

Each capability should be designed according to the METR Standard, which requires the following Python format:
\begin{verbatim}
class Capability:
    @staticmethod
    def repr_tasks() -> dict[str, dict]:
        # Returns a mapping from task ids to arbitrary data that is
        # specific to each capability.
        # Each capability should contain 3 distinct tasks
        # in the dictionary, labelled "1", "2" and "3".
        # Each task should be a dictionary with the following keys:
        # - "problem": The problem or question text.
        # - "answer": The expected answer to the problem.
        # ... implementation ...

    @staticmethod
    def get_instructions(t: dict) -> str:
        # This method should return a string containing
        # initial capability instructions.
        # ... implementation ...

    @staticmethod
    def score(t: dict, submission: str) -> float | None:
        # ‘score‘ is called after the subject model submits a solution.
        # Return 0 on failure and 1 on success.
        # Throwing an exception will result in a score of 0.
        # ... implementation ...
\end{verbatim}

The following points should be considered while designing the Capability class:
\begin{enumerate}
    \item The instructions should enforce the final answer in the form "ANSWER: \$ANSWER" ONLY for those capabilities that expect a short and objective answer. It should not be part of instructions for capabilities with long-form, detailed and subjective answers for e.g. explanations, proofs, essays, etc.
    \item If the instructions enforces the final answer in the form "ANSWER: \$ANSWER", then the score function should first call a helper function to parse the submission string and extract the answer:
\begin{verbatim}
def parse_submission(submission: str) -> str:
    # Parse the submission string to extract the answer
    # based on the "ANSWER" keyword.
    # Return an empty string if no match is found.
\end{verbatim}
    \item The score function should use a helper function that uses LLM as a judge to score the submission:
\begin{verbatim}
def evaluate_with_llm_judge(
    submission: str,
    target: str,
) -> bool:
    # Evaluate the submission using an LLM judge.
\end{verbatim}
    \item DO NOT re-implement the \texttt{parse\_submission()} or \texttt{evaluate\_with\_llm\_judge()} helper functions.
\end{enumerate}

Respond precisely in the following format, including the JSON start and end markers: \\

THOUGHT: <THOUGHT> \\

RESPONSE JSON:
\begin{verbatim}
{
    "capability_0": <JSON>,
    "capability_1": <JSON>,
    ...
}
\end{verbatim}

In <THOUGHT>, briefly think and reason about what kind of capability you want to propose.
In <JSON>, provide a JSON response of the new capability with the following fields:
\begin{itemize}
    \item[-] "name": A concise, descriptive label (lowercase, no spaces, e.g., \texttt{math\_competition\_algebra}).
    \item[-] "description": A clear explanation of what the capability entails (e.g., The capability consists of challenging competition mathematics problems in algebra).
    \item[-] "domain": The domain to which the capability belongs to (e.g., math, physics, etc.).
    \item[-] "class": The fully implemented Python code for the Capability class. This should be easily human-readable.
\end{itemize}

Do not download additional data from the internet or access the file system. \\

Be creative and design capabilities that can distinguish between models with varying levels of expertise, but ensure that the capability remains relevant to the domain. Also ensure that the proposed capabilities ARE DISTINCT compared to the existing capabilities. Names of all existing capabilities will be provided. \\

Your response will be automatically parsed so ensure it adheres to the specified format.
\end{tcolorbox}

\begin{tcolorbox}[
  colback=promptheaderbg,       
  colframe=promptheaderbg,       
  width=\linewidth,     
  boxrule=0.5pt,        
  boxsep=0pt,
  left=5pt, right=5pt, top=5pt, bottom=5pt,
  arc=4pt, 
  before skip=10pt,
  after skip=0pt,
  enhanced,
  sharp corners=south,        
  listing only,         
  listing options={basicstyle=\ttfamily\small}
]
\color{white}\textbf{Capability Generation User Prompt}
\end{tcolorbox}
\begin{tcolorbox}[
  colback=promptframebg,       
  colframe=promptheaderbg,       
  width=\linewidth,     
  boxrule=0.5pt,        
  boxsep=0pt,
  left=5pt, right=5pt, top=5pt, bottom=5pt,
  arc=0pt,
  before skip=0pt,
  after skip=10pt,
  enhanced,
  sharp corners,        
  listing only,         
  listing options={basicstyle=\ttfamily\small}
]
A sample capability JSON is provided below. The names of all existing capabilities are also provided. \\

Sample capability:\newline
\texttt{sample\_capability\_json} \\

Existing capability names:\newline
\texttt{prev\_capabilities} \\

Generate \texttt{num\_gen\_capabilities} new, interesting capabilities for the \texttt{"capability\_area"} area within the \texttt{domain} domain.
\end{tcolorbox}

\clearpage

\subsection{Task Generation Prompts}\label{sec:task_gen_prompts}

\begin{tcolorbox}[
  colback=promptheaderbg,       
  colframe=promptheaderbg,       
  width=\linewidth,     
  boxrule=0.5pt,        
  boxsep=0pt,
  left=5pt, right=5pt, top=5pt, bottom=5pt,
  arc=4pt, 
  before skip=10pt,
  after skip=0pt,
  enhanced,
  sharp corners=south,        
  listing only,         
  listing options={basicstyle=\ttfamily\small}
]
\color{white}\textbf{Task Generation System Prompt}
\end{tcolorbox}
\begin{tcolorbox}[
  colback=promptframebg,       
  colframe=promptheaderbg,       
  width=\linewidth,     
  boxrule=0.5pt,        
  boxsep=0pt,
  left=5pt, right=5pt, top=5pt, bottom=5pt,
  arc=0pt,
  before skip=0pt,
  after skip=10pt,
  enhanced,
  sharp corners,        
  listing only,         
  listing options={basicstyle=\ttfamily\small}
]
You are an expert in designing tasks for a given capability. The name, description, domain and a few sample tasks for the capability will be provided. You will be particularly rewarded for designing diverse tasks spanning a wide range of difficulty levels for the given capability. \\

Respond precisely in the following format, including the JSON start and end markers: \\

THOUGHT: <THOUGHT>\newline
RESPONSE JSON:
\begin{verbatim}
{
    "task_1": <STR>,
    "task_2": <STR>,
    ...
}
\end{verbatim}

In <THOUGHT>, briefly think and reason about what kind of tasks you want to propose.\newline
In <STR>, provide a string containing the task text. \\

Be careful to make sure that all proposed tasks are unique. Also ensure that all tasks are within the scope of the given capability.
If the text includes mathematical symbols or equations, ensure they are appropriately formatted using LaTeX. Ensure the single backlash "$\backslash$" included in a LateX string is escaped as "$\backslash$$\backslash$". For example, the LaTeX string "$\backslash[2x + 3 = 11\backslash]$" should be formatted as "$\backslash\backslash[2x + 3 = 11\backslash\backslash]$" in the task text. \\

Your response will be automatically parsed so ensure it adheres to the specified format.
\end{tcolorbox}

\begin{tcolorbox}[
  colback=promptheaderbg,       
  colframe=promptheaderbg,       
  width=\linewidth,     
  boxrule=0.5pt,        
  boxsep=0pt,
  left=5pt, right=5pt, top=5pt, bottom=5pt,
  arc=4pt, 
  before skip=10pt,
  after skip=0pt,
  enhanced,
  sharp corners=south,        
  listing only,         
  listing options={basicstyle=\ttfamily\small}
]
\color{white}\textbf{Task Generation User Prompt}
\end{tcolorbox}
\begin{tcolorbox}[
  colback=promptframebg,       
  colframe=promptheaderbg,       
  width=\linewidth,     
  boxrule=0.5pt,        
  boxsep=0pt,
  left=5pt, right=5pt, top=5pt, bottom=5pt,
  arc=0pt,
  before skip=0pt,
  after skip=10pt,
  enhanced,
  sharp corners,        
  listing only,         
  listing options={basicstyle=\ttfamily\small}
]
Design tasks for the following capability: \\

Name: \texttt{capability\_name}\newline
Description: \texttt{capability\_description}\newline
Domain: \texttt{capability\_domain}\newline
Sample tasks:\newline
\texttt{capability\_sample\_tasks} \\

Generate \texttt{num\_gen\_tasks} new tasks for the given capability.
\end{tcolorbox}

\begin{tcolorbox}[
  colback=promptheaderbg,       
  colframe=promptheaderbg,       
  width=\linewidth,     
  boxrule=0.5pt,        
  boxsep=0pt,
  left=5pt, right=5pt, top=5pt, bottom=5pt,
  arc=4pt, 
  before skip=10pt,
  after skip=0pt,
  enhanced,
  sharp corners=south,        
  listing only,         
  listing options={basicstyle=\ttfamily\small}
]
\color{white}\textbf{Task Solver System Prompt}
\end{tcolorbox}
\begin{tcolorbox}[
  colback=promptframebg,       
  colframe=promptheaderbg,       
  width=\linewidth,     
  boxrule=0.5pt,        
  boxsep=0pt,
  left=5pt, right=5pt, top=5pt, bottom=5pt,
  arc=0pt,
  before skip=0pt,
  after skip=10pt,
  enhanced,
  sharp corners,        
  listing only,         
  listing options={basicstyle=\ttfamily\small}
]
You are an expert in completing tasks for the \texttt{capability\_name} capability in the \texttt{capability\_domain} domain. Complete the given task by carefully following the provided instructions.
\end{tcolorbox}

\clearpage

\begin{tcolorbox}[
  colback=promptheaderbg,       
  colframe=promptheaderbg,       
  width=\linewidth,     
  boxrule=0.5pt,        
  boxsep=0pt,
  left=5pt, right=5pt, top=5pt, bottom=5pt,
  arc=4pt, 
  before skip=10pt,
  after skip=0pt,
  enhanced,
  sharp corners=south,        
  listing only,         
  listing options={basicstyle=\ttfamily\small}
]
\color{white}\textbf{Task Verifier System Prompt}
\end{tcolorbox}
\begin{tcolorbox}[
  colback=promptframebg,       
  colframe=promptheaderbg,       
  width=\linewidth,     
  boxrule=0.5pt,        
  boxsep=0pt,
  left=5pt, right=5pt, top=5pt, bottom=5pt,
  arc=0pt,
  before skip=0pt,
  after skip=10pt,
  enhanced,
  sharp corners,        
  listing only,         
  listing options={basicstyle=\ttfamily\small}
]
You are an expert in evaluating answers to problems for the \texttt{capability\_domain} domain. Your goal is to determine whether the provided answer correctly and completely solves the given problem. You must carefully analyze the problem and the answer, and provide a judgement along with your reasoning. \\

Respond precisely in the following format: \\

THOUGHT: <THOUGHT>\newline
JUDGEMENT:\newline
<JUDGEMENT> \\

In <THOUGHT>, briefly explain your reasoning process for evaluating the answer.\newline
In <JUDGEMENT>, respond with "yes" if the answer correctly and completely solves the problem, otherwise respond with "no". \\

Be objective and thorough in your evaluation. Ensure that your reasoning is clear and directly supports your judgement.
\end{tcolorbox}

\begin{tcolorbox}[
  colback=promptheaderbg,       
  colframe=promptheaderbg,       
  width=\linewidth,     
  boxrule=0.5pt,        
  boxsep=0pt,
  left=5pt, right=5pt, top=5pt, bottom=5pt,
  arc=4pt, 
  before skip=10pt,
  after skip=0pt,
  enhanced,
  sharp corners=south,        
  listing only,         
  listing options={basicstyle=\ttfamily\small}
]
\color{white}\textbf{Task Verifier User Prompt}
\end{tcolorbox}
\begin{tcolorbox}[
  colback=promptframebg,       
  colframe=promptheaderbg,       
  width=\linewidth,     
  boxrule=0.5pt,        
  boxsep=0pt,
  left=5pt, right=5pt, top=5pt, bottom=5pt,
  arc=0pt,
  before skip=0pt,
  after skip=10pt,
  enhanced,
  sharp corners,        
  listing only,         
  listing options={basicstyle=\ttfamily\small}
]
Evaluate the following problem and answer for the \texttt{capability\_name} capability in the \texttt{capability\_domain} domain: \\

Problem: {problem}\newline
Answer: {answer} \\

Determine if the answer correctly and completely solves the problem. Provide your reasoning and judgement.
\end{tcolorbox}

\subsection{Static Dataset Comparison Prompt}
\label{sec:static_dataset_categorization}

\begin{tcolorbox}[
  colback=promptheaderbg,       
  colframe=promptheaderbg,       
  width=\linewidth,     
  boxrule=0.5pt,        
  boxsep=0pt,
  left=5pt, right=5pt, top=5pt, bottom=5pt,
  arc=4pt, 
  before skip=10pt,
  after skip=0pt,
  enhanced,
  sharp corners=south,        
  listing only,         
  listing options={basicstyle=\ttfamily\small}
]
\color{white}\textbf{Task Verifier System Prompt}
\end{tcolorbox}
\begin{tcolorbox}[
  colback=promptframebg,       
  colframe=promptheaderbg,       
  width=\linewidth,     
  boxrule=0.5pt,        
  boxsep=0pt,
  left=5pt, right=5pt, top=5pt, bottom=5pt,
  arc=0pt,
  before skip=0pt,
  after skip=10pt,
  enhanced,
  sharp corners,        
  listing only,         
  listing options={basicstyle=\ttfamily\small}
]
Available mathematical \textsf{areas}:\\

\texttt{{area\_1}}\\
\texttt{{area\_2}}\\
\texttt{{...}}\\

Problem:
\texttt{{question}}\\

Answer with ONLY the exact \textsf{area name} from the list above, or \texttt{none} if the problem does not fit any of the given \textsf{areas}.
\end{tcolorbox}

\begin{tcolorbox}[
  colback=promptheaderbg,       
  colframe=promptheaderbg,       
  width=\linewidth,     
  boxrule=0.5pt,        
  boxsep=0pt,
  left=5pt, right=5pt, top=5pt, bottom=5pt,
  arc=4pt, 
  before skip=10pt,
  after skip=0pt,
  enhanced,
  sharp corners=south,        
  listing only,         
  listing options={basicstyle=\ttfamily\small}
]
\color{white}\textbf{Task Verifier User Prompt}
\end{tcolorbox}
\begin{tcolorbox}[
  colback=promptframebg,       
  colframe=promptheaderbg,       
  width=\linewidth,     
  boxrule=0.5pt,        
  boxsep=0pt,
  left=5pt, right=5pt, top=5pt, bottom=5pt,
  arc=0pt,
  before skip=0pt,
  after skip=10pt,
  enhanced,
  sharp corners,        
  listing only,         
  listing options={basicstyle=\ttfamily\small}
]
You are an expert in mathematical problem categorization. \\

Given the list of \textsf{area names} and a math word problem, respond with ONLY the exact \textsf{area name} from the list. If the problem does not fit any of the given \textsf{areas}, respond with \texttt{none}. No explanations, no extra text.
\end{tcolorbox}

\end{document}